\pdfoutput=1
\documentclass[10pt,journal,cspaper,compsoc]{IEEEtran}
\ifCLASSOPTIONcompsoc
\else
\fi
\ifCLASSINFOpdf
\else
\fi
\usepackage[cmex10]{amsmath}
\usepackage{amssymb,amsthm}
\usepackage{algorithm}
\usepackage{algorithmicx}
\usepackage{algpseudocode}
\usepackage{url}

\usepackage{flushend}
\usepackage{tikz}
\usepackage{pgfplots}
\usetikzlibrary{plotmarks}
\pgfplotsset{compat=newest}

\usetikzlibrary{trees,shapes.geometric,arrows,positioning}

\usepackage{booktabs} 
\usepackage{multirow} 
\usepackage{pifont}   
\usepackage{paralist, tabularx}

\newcommand*{\etal}{\textit{et al.}\ }
\DeclareMathOperator*{\argmax}{arg\,max}

\newcommand{\showodsf}[2]{%
\mbox{\input{data/pr_curves/#1_#2_ods_f.txt}\hspace{-2.5pt}}%
}

\newcommand{\showodsfb}[2]{%
\mbox{\input{data/segm_bsds/test_#1_#2_ods_f.txt}\hspace{-2.5pt}}%
}

\definecolor{mygreen}{RGB}{69,182,73}

\begin{document}
%
\title{Multiscale Combinatorial Grouping\\for Image Segmentation and\\Object Proposal Generation}
%
%
%
%

\author{Jordi Pont-Tuset*, 
\and
Pablo Arbel\'aez*, 
\and
Jonathan T. Barron,~\IEEEmembership{Member,~IEEE,}\\
\and
Ferran Marques,~\IEEEmembership{Senior Member,~IEEE,}
\and
Jitendra Malik,~\IEEEmembership{Fellow,~IEEE}
\IEEEcompsocitemizethanks{\IEEEcompsocthanksitem J. Pont-Tuset and F. Marques are
with the Department of Signal Theory and Communications, Universitat Polit\`{e}cnica de Catalunya, BarcelonaTech (UPC), Spain.
E-mail: \{jordi.pont,ferran.marques\}@upc.edu
\IEEEcompsocthanksitem P. Arbel\'aez is
with the Department of Biomedical Engineering, Universidad de los Andes, Colombia.
E-mail: pa.arbelaez@uniandes.edu.co
\IEEEcompsocthanksitem J. T. Barron, and J. Malik are with the Department of Electrical Engineering and Computer Science,
University of California at Berkeley, Berkeley, CA 94720.
E-mail: \{barron,malik\}@eecs.berkeley.edu\protect\\
}
\thanks{* The first two authors contributed equally}}

%
%

\markboth{}%
{}
%


\IEEEcompsoctitleabstractindextext{%
\begin{abstract}
We propose a unified approach for bottom-up hierarchical image segmentation and object proposal generation for recognition, called Multiscale Combinatorial Grouping (MCG). 
For this purpose, we first develop a fast normalized cuts algorithm. We then propose a high-performance hierarchical segmenter that makes effective use of multiscale information. 
Finally, we propose a grouping strategy that combines our multiscale regions into highly-accurate object proposals by exploring efficiently their combinatorial space.
We also present Single-scale Combinatorial Grouping (SCG), a faster version of MCG that produces
competitive proposals in under five second per image.
We conduct an extensive and comprehensive empirical validation on the BSDS500, SegVOC12, SBD, and COCO datasets, showing that MCG produces state-of-the-art contours, hierarchical regions, and object proposals.
\end{abstract}


\begin{keywords}
Image segmentation, object proposals, normalized cuts.
\end{keywords}}

\maketitle

\IEEEdisplaynotcompsoctitleabstractindextext

%
\IEEEpeerreviewmaketitle

\section{Introduction}
\IEEEPARstart{T}{wo} paradigms have shaped the field of object recognition in the last decade. The first one, popularized by the Viola-Jones face detection algorithm~\cite{viola_jones:IJCV04}, formulates object localization as window classification. 
The basic scanning-window architecture, relying on histograms of gradients and linear support vector machines, was introduced by Dalal and Triggs \cite{dalal_triggs:CVPR05} in the context of pedestrian detection and is still at the core of seminal object detectors on the PASCAL challenge such as Deformable Part Models \cite{FelzenszwalbTPAMI2010}.

The second paradigm relies on perceptual grouping to provide a limited number of high-quality and category-independent object proposals, which can then be described with richer representations and used as input to more sophisticated learning methods.
Examples in this family are~\cite{Malisiewicz2007,Gu_etal:cvpr2009}. Recently, this approach has dominated the PASCAL segmentation challenge \cite{Carreira2012,Ion2014,Arbelaez2012,o2p}, improved object detection~\cite{Girshick:cvpr2014}, fine-grained categorization \cite{Zhang2014} and proven competitive in large-scale classification \cite{Uijlings2013}.

Since the power of this second paradigm is critically dependent on the accuracy and the number of object proposals, an increasing body of research has delved into the problem of their generation \cite{Kraehenbuehl2014,Rantalankila2014,Humayun2014,Uijlings2013,Alexe2012,Kim2012,Carreira2012b,Endres2014}.
However, those approaches typically focus on learning generic properties of objects from a set of examples, while reasoning on a fixed set of regions and contours produced by external bottom-up segmenters such as \cite{Arbelaez2011,Felzenszwalb2004}. 

\begin{figure}[t]
\begin{center}
\setlength{\fboxsep}{0pt}
 \fbox{\includegraphics[width=0.32\linewidth]{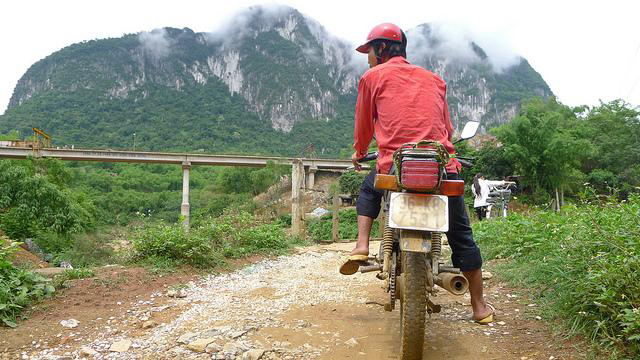}}
 \hfill
 \fbox{\includegraphics[width=0.32\linewidth]{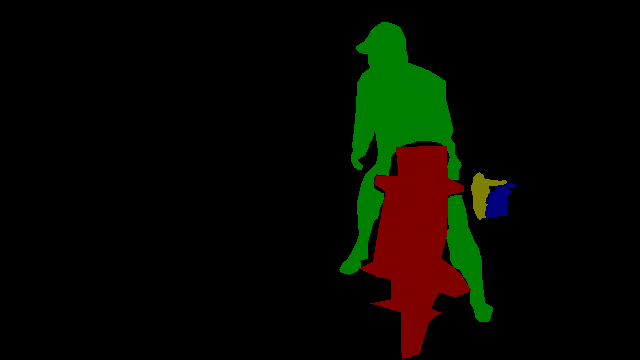}}
\hfill
 \fbox{\includegraphics[width=0.32\linewidth]{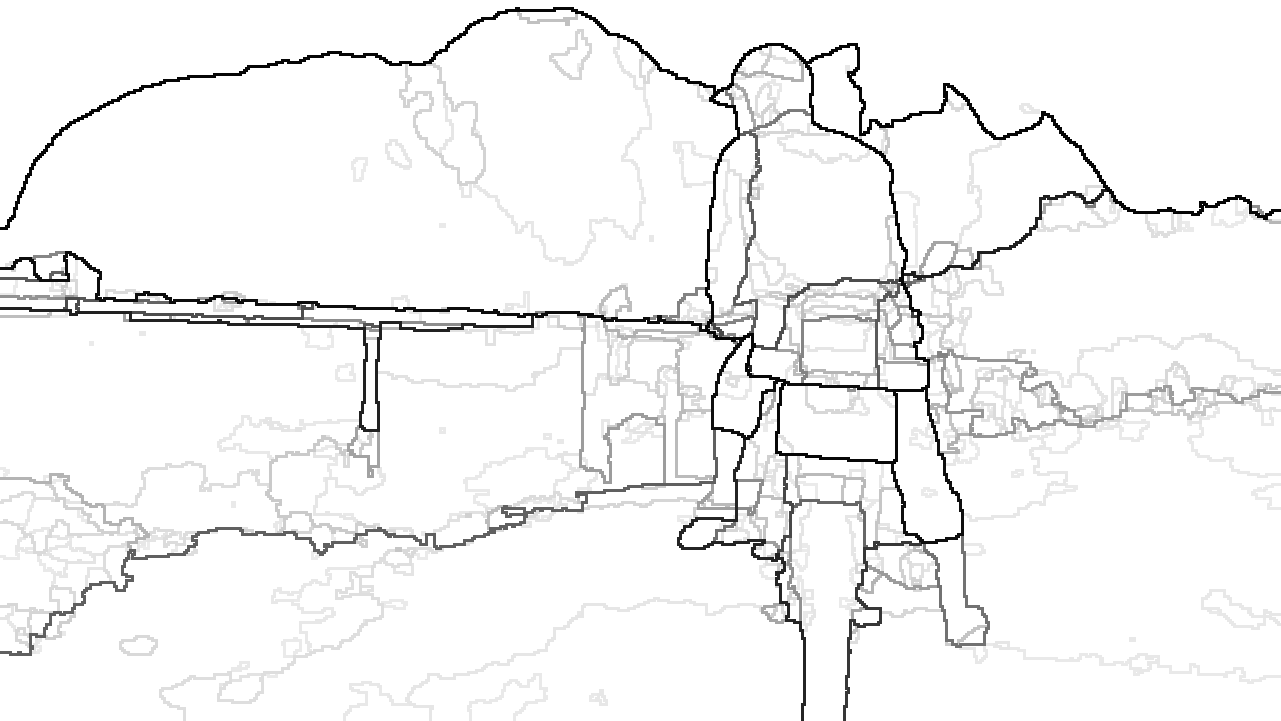}}\\[2mm]
 \fbox{\includegraphics[width=0.32\linewidth]{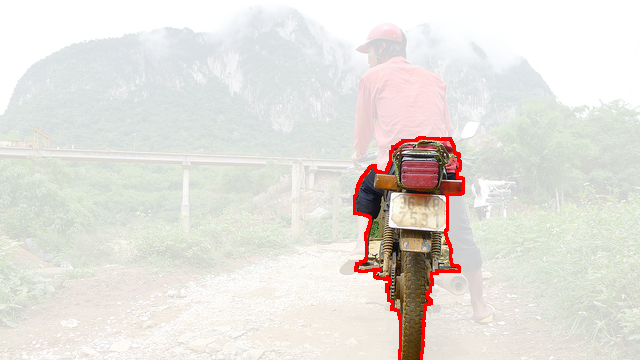}}
 \hfill
 \fbox{\includegraphics[width=0.32\linewidth]{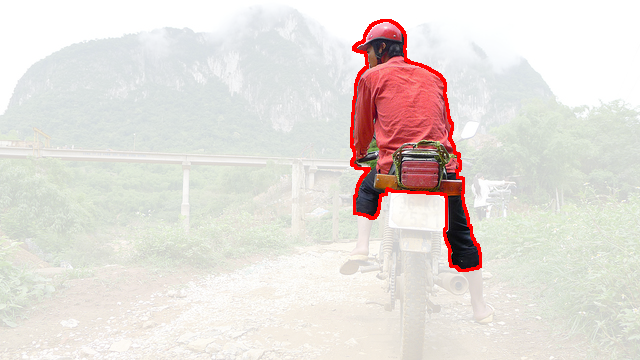}}
\hfill
 \fbox{\includegraphics[width=0.32\linewidth]{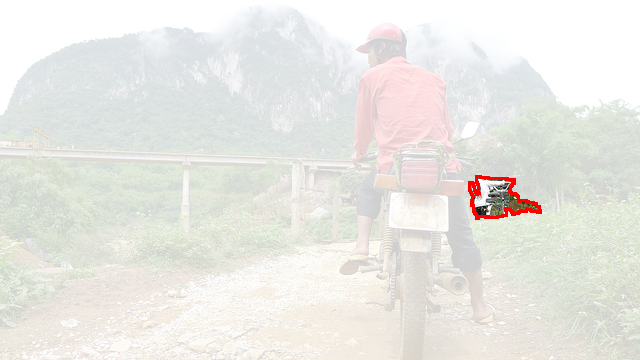}}
\end{center}
\vspace{-1mm}
   \caption{\textbf{Top:} original image, instance-level ground truth from COCO and our multiscale hierarchical segmentation. \textbf{Bottom:} our best object proposals among 150.}
\label{fig1}
\end{figure}

\begin{figure*}[t]
\begin{center}
   \includegraphics[width=0.9\linewidth]{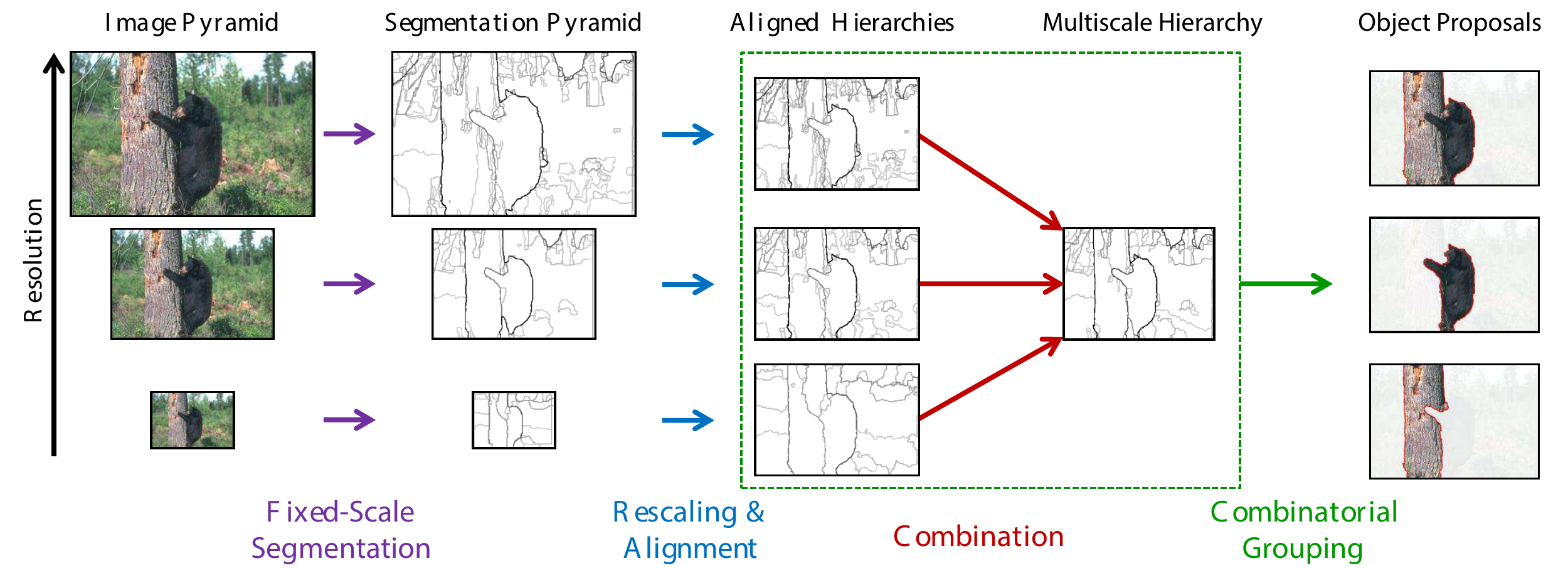}
\end{center}
   \caption{\textbf{Multiscale Combinatorial Grouping}. Starting from a multiresolution image pyramid, we perform hierarchical segmentation at each scale independently. 
We align these multiple hierarchies and combine them into a single multiscale segmentation hierarchy. 
Our grouping component then produces a ranked list of object proposals by efficiently exploring the combinatorial space of these regions.}
\label{fig:overview}
\end{figure*}

In this paper, we propose a unified approach to multiscale hierarchical segmentation and object proposal generation called Multiscale Combinatorial Grouping (MCG). 
Fig.~\ref{fig1} shows an example of our results and Fig.~\ref{fig:overview} an overview of our pipeline. Our main contributions are:
\begin{itemize}
\item An efficient normalized cuts algorithm, which in practice provides a $20\times$ speed-up to the eigenvector computation required for contour globalization \cite{Arbelaez2011,renNIPS12} (Sect.~\ref{sec:fast_eigen}).
\item A state-of-the-art hierarchical segmenter that leverages multiscale information (Sect.~\ref{sect:multi}).
\item A grouping algorithm that produces accurate object proposals by efficiently exploring the combinatorial space of our multiscale regions (Sect.~\ref{sec:obj_prop}).
\end{itemize}
We conduct a comprehensive and large-scale empirical validation.
On the BSDS500 (Sect.~\ref{sect:bsds}) we report significant progress in contour detection and hierarchical segmentation. 
On the VOC2012, SBD, and COCO segmentation datasets (Sect.~\ref{sect:experim_prop}), our proposals obtain overall state-of-the-art accuracy
both as segmented proposals and as bounding boxes.
MCG is efficient, its good generalization power makes it parameter free in practice, and it provides a ranked set of proposals that are competitive in all regimes of number of proposals.


\section{Related Work}
\label{sec:related_work}

For space reasons, we focus our review on recent normalized cut algorithms and object proposals for recognition.

\paragraph*{\textbf{Fast normalized cuts}} The efficient computation of normalized-cuts eigenvectors has been the subject of recent work, as it is often the computational bottleneck in grouping algorithms. 
Taylor \cite{TaylorCVPR13} presented a technique for using a simple watershed oversegmentation to reduce the size of the eigenvector problem, sacrificing accuracy for speed. 
We take a similar approach of solving the eigenvector problem in a reduced space, though we use simple image-pyramid operations on the affinity matrix (instead of a separate segmentation algorithm) and we see no loss in performance despite a 20$\times$ speed improvement. 
Maire and Yu \cite{MaireICCV2013} presented a novel multigrid solver for producing eigenvectors at multiple scales, which speeds up fine-scale eigenvector computation by leveraging coarse-scale solutions. 
Our technique also uses the scale-space structure of an image, but instead of solving the problem at multiple scales, we simply reduce the scale of the problem, solve it at a reduced scale, and then upsample the solution while preserving the structure of the image. 
As such, our technique is faster and much simpler, requiring only a few lines of code wrapped around a standard sparse eigensolver.

\paragraph*{\textbf{Object Proposals}} Class-independent methods that generate object hypotheses can be divided into those whose output is an image window
and those that generate segmented proposals.

Among the former, Alexe \etal \cite{Alexe2012} propose an \textit{objectness} measure to score randomly-sampled image
windows based on low-level features computed on the superpixels of~\cite{Felzenszwalb2004}.
%
Manen \etal \cite{Manen2013} propose to use the Randomized Prim's algorithm, Zitnick \etal \cite{Zitnick2014} group contours directly to produce object windows,
and Cheng \etal \cite{Cheng2014} generate box proposals at 300 images per second.   
In contrast to these approaches, we focus on the finer-grained task of pixel-accurate object extraction, rather than on window selection. However, by just taking the bounding box around our segmented proposals, our results
 are also state of the art as window proposals.  

Among the methods that produce segmented proposals, Carreira and Sminchisescu~\cite{Carreira2012b} hypothesize a set of placements of fore- and background seeds and,
for each configuration, solve a constrained parametric min-cut (CPMC) problem to generate a pool of object hypotheses.
Endres and Hoiem~\cite{Endres2014} base their category-independent object proposals on an iterative generation of a hierarchy of regions,
based on the contour detector of~\cite{Arbelaez2011} and occlusion boundaries of~\cite{Hoiem2011}.
Kim and Grauman~\cite{Kim2012} propose to match parts of the shape of exemplar objects, regardless of their class,
to detected contours by~\cite{Arbelaez2011}. 
They infer the presence and shape of a proposal object by adapting the matched object to the computed superpixels.

Uijlings \etal \cite{Uijlings2013} present a selective search algorithm based on segmentation.
Starting with the superpixels of~\cite{Felzenszwalb2004} for a variety of color spaces,
they produce a set of segmentation hierarchies by region merging, which are used
to produce a set of object proposals. 
While we also take advantage of different hierarchies to gain diversity, we leverage multiscale information rather than different color spaces.

Recently, two works proposed to train a cascade of classifiers to learn which sets of regions should be merged to form objects.
Ren and Shankhnarovich~\cite{Ren_2013_CVPR} produce full region hierarchies by iteratively merging pairs of regions and adapting the classifiers
to different scales. Weiss and Taskar~\cite{weiss2013scalpel} specialize the classifiers also to size and class of the annotated
instances to produce object proposals.

Malisiewicz and Efros~\cite{Malisiewicz2007} took one of the first steps towards combinatorial grouping, by running multiple segmenters with different parameters and merging up to three adjacent regions.
In~\cite{Arbelaez2012}, another step was taken by considering hierarchical segmentations at three different scales and combining pairs and triplets of adjacent regions from the two coarser scales to produce object proposals.

The most recent wave of object proposal algorithms is represented by \cite{Kraehenbuehl2014}, \cite{Rantalankila2014}, and~\cite{Humayun2014}, which all keep
the quality of the seminal proposal works while improving the speed considerably.
Kr\"ahenb\"uhl and Koltun~\cite{Kraehenbuehl2014} find object proposal by identifying critical level sets in
geodesic distance transforms, based on seeds placed in learnt places in the image.
Rantalankila \etal \cite{Rantalankila2014} perform a global and local search in the space of sets of superpixels.
Humayun \etal \cite{Humayun2014} reuse a graph to perform many parametric min-cuts over different seeds in order to speed the process up.



A substantial difference between our approach and previous work is that, instead of relying on pre-computed hierarchies or superpixels, we propose a unified approach that produces and groups high-quality multiscale regions.
With respect to the combinatorial approaches of \cite{Malisiewicz2007,Arbelaez2012}, our main contribution is to develop efficient algorithms to explore a much larger combinatorial space by taking into account a set of object examples, increasing thus the likelihood of having complete objects in the pool of proposals. 
Our approach has therefore the flexibility to adapt to specific applications and types of objects, and can produce proposals at any trade-off between their number and their accuracy.

\section{The Segmentation Algorithm}
Consider a segmentation of the image into regions that partition its domain $\mathcal{S} = \{ S_i\}_i$. 
A segmentation hierarchy is a family of partitions $\{\mathcal{S}^*,\mathcal{S}^{1},...,\mathcal{S}^{L}\}$ such that: (1) $\mathcal{S}^{*}$ is the finest set of \emph{superpixels}, (2) $\mathcal{S}^{L}$ is the complete domain, and (3) regions from coarse levels are unions of regions from fine levels.
A hierarchy where each level $\mathcal{S}^{i}$ is assigned a real-valued index $\lambda_i$ can be represented by a dendrogram, a region tree where the height of each node is its index. 
Furthermore, it can also be represented as an ultrametric contour map (UCM), an image obtained by weighting the boundary of each pair of adjacent regions in the hierarchy by the index at which they are merged \cite{Najman1996,Arbelaez:POCV06}. 
This representation unifies the problems of contour detection and hierarchical image segmentation: a threshold at level $\lambda_i$ in the UCM produces the segmentation $\mathcal{S}^i$. 

\begin{figure}[t]
\includegraphics[width=\linewidth]{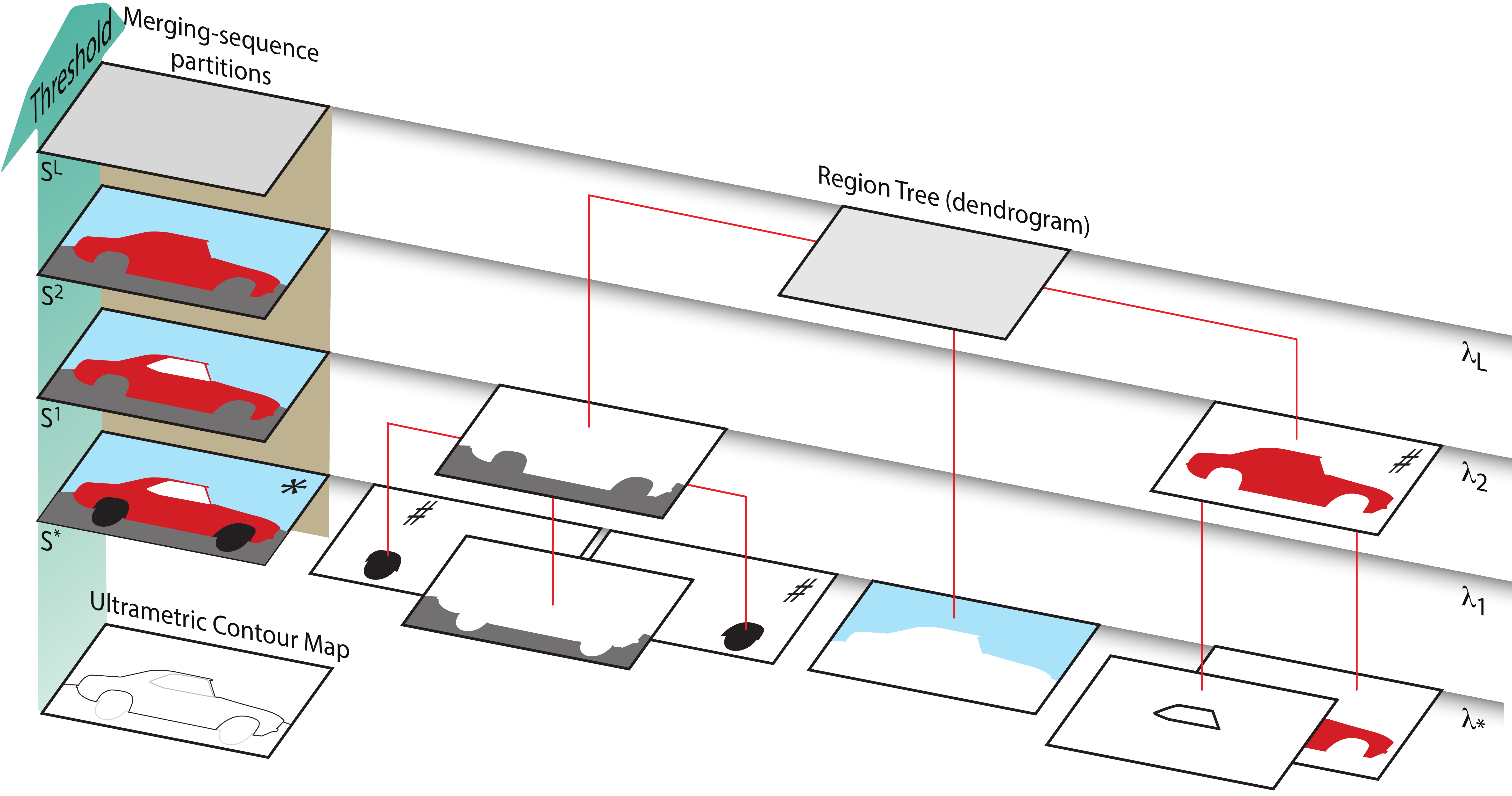}
\caption{\textbf{Duality between a UCM and a region tree}: Schematic view of the dual representation of a segmentation hierarchy
as a region dendrogram and as an ultrametric contour map.}
\label{fig:hierarchy}
\end{figure}

Figure~\ref{fig:hierarchy} schematizes these concepts.
First, the lower left corner shows the probability of boundary of a UCM.
One of the main properties of a UCM is that when we threshold the contour strength at a certain value,
we obtain a closed boundary map, and thus a partition.
Thresholding at different $\lambda_i$, therefore, we obtain the so-called merging-sequence partitions
(left column in Figure~\ref{fig:hierarchy}); named after the fact that a step in this sequence
corresponds to merging the set of regions sharing the boundary of strength exactly $\lambda_i$.

For instance, the boundary between the wheels and the floor has strength $\lambda_1$,
thus thresholding the contour above $\lambda_1$ makes the wheels \textit{merge} with the floor.
If we represent the regions in a partition as nodes of a graph, we can then represent the result
of merging them as their parent in a tree.
The result of sweeping all $\lambda_i$ values can therefore be represented as a region tree, 
whose root is the region representing the whole image (right part of Figure~\ref{fig:hierarchy}).
Given that each \textit{merging} is associated with a contour strength,
the region tree is in fact a region dendogram.

As an example, in the gPb-ucm algorithm of~\cite{Arbelaez2011}, brightness, color and texture gradients at three fixed disk sizes are first computed. 
These local contour cues are globalized using spectral graph-partitioning, resulting in the gPb contour detector. 
Hierarchical segmentation is then performed by iteratively merging adjacent regions based on the average gPb strength on their common boundary. 
This algorithm produces therefore a tree of regions at multiple levels of homogeneity in brightness, color and texture, and the boundary strength of its UCM can be interpreted as a measure of contrast. 

Coarse-to-fine is a powerful processing strategy in computer vision. 
We exploit it in two different ways to develop an efficient, scalable and high-performance segmentation algorithm: 
(1) To speed-up spectral graph partitioning and (2) To create aligned segmentation hierarchies.

\subsection{Fast Downsampled Eigenvector Computation}
\label{sec:fast_eigen}

The normalized cuts criterion is a key globalization mechanism of recent high-performance contour detectors such as \cite{Arbelaez2011, renNIPS12}.
Although powerful, such spectral graph partitioning has a significant computational cost and memory footprint that limit its scalability. 
In this section, we present an efficient normalized cuts approximation which in practice preserves full performance for contour detection, while having low memory requirements and providing a $20\times$ speed-up.

Given a symmetric affinity matrix $A$, we would like to compute the $k$ smallest eigenvectors of the Laplacian of $A$. 
Directly computing such eigenvectors can be very costly even with sophisticated solvers, due to the large size of $A$. 
We therefore present a technique for efficiently approximating the eigenvector computation by taking advantage of the multiscale nature of our problem: 
$A$ models affinities between pixels in an image, and images naturally lend themselves to multiscale or pyramid-like representations and algorithms.

Our algorithm is inspired by two observations: 1) if $A$ is bistochastic (the rows and columns of $A$ sum to 1) then the eigenvectors of the Laplacian $A$ are equal to the eigenvectors of the Laplacian of $A^2$, and 2) because of the scale-similar nature of images, the eigenvectors of a ``downsampled'' version of $A$ in which every other pixel has been removed should be similar to the eigenvectors of $A$.
Let us define ${\tt pixel\_decimate}\left( A \right)$, which takes an affinity matrix $A$ and returns the set of indices of rows/columns in $A$ corresponding to a decimated version of the image from which $A$ was constructed.
That is, if $i = {\tt pixel\_decimate}\left( A \right)$, then $A\left[i,i\right]$ is a decimated matrix in which alternating rows and columns \emph{of the image} have been removed. 
Computing the eigenvectors of $A\left[i,i\right]$ works poorly, as decimation disconnects pixels in the affinity matrix, but the eigenvectors of the decimated squared affinity matrix $A^2 \left[i,i\right]$ are similar to those of $A$, because by squaring the matrix before decimation we intuitively allow each pixel to propagate information to all of its neighbors in the graph, maintaining connections even after decimation.
Our algorithm works by efficiently computing $A^2 \left[i,i\right]$ as $A\left[:,i\right]^\mathrm{T}A\left[:,i\right]$\footnote{The \textit{Matlab-like} notation $A\left[:,i\right]$ indicates keeping the columns of matrix $A$ whose indices are in the set $i$.} (the naive approach of first squaring $A$ and then decimating it is prohibitively expensive), computing the eigenvectors of $A^2 \left[i,i\right]$, and then ``upsampling'' those eigenvectors back to the space of the original image by pre-multiplying by $A\left[:,i\right]$.
This squaring-and-decimation procedure can be applied recursively several times, each application improving efficiency while slightly sacrificing accuracy. 


\begin{algorithm}
\caption{${\tt dncuts}(A, d, k)$}  
\begin{algorithmic}[1]
\State $A_0 \gets A$
\For{$s = \left[ 1, 2, \dots, d \right]$}
\State $i_s \gets {\tt pixel\_decimate}\left( A_{s-1} \right)$
\State $B_s \gets A_{s-1}\left[\,:\,,\, i_s \,\right]$
\State $C_s \gets {\tt diag}( B_s \vec{1} )^{-1} B_s$
\State $A_s \gets C_s^\mathrm{T} B_s$
\EndFor
\State $X_d \gets {\tt ncuts}(A_d, k)$
\For{$s = \left[ d, d-1, \dots, 1\right]$} 
\State $X_{s-1} \gets C_s X_s$
\EndFor
\State \Return ${\tt whiten}(X_0)$
\end{algorithmic}
\label{alg:dncuts}
\end{algorithm}


Pseudocode for our algorithm, which we call ``DNCuts'' (Downsampled Normalized Cuts) is given in Algorithm~\ref{alg:dncuts}, where $A$ is our affinity matrix and $d$ is the number of times that our squaring-and-decimation operation is applied.
Our algorithm repeatedly applies our joint squaring-and-decimation procedure, computes the smallest $k$ eigenvectors of the final ``downsampled'' matrix $A_d$ by using a standard sparse eigensolver ${\tt ncuts}(A_d, k)$, and repeatedly ``upsamples'' those eigenvectors.
Because our $A$ is not bistochastic and  decimation is not an orthonormal operation, we must do some normalization throughout the algorithm (line $5$) and whiten the resulting eigenvectors (line $10$).
We found that values of $d=2$ or $d=3$ worked well in practice.
Larger values of $d$ yielded little speed improvement (as much of the cost is spent downsampling $A_0$) and start negatively affecting accuracy. 
Our technique is similar to Nystrom's method for computing the eigenvectors of a subset of $A$, but our squaring-and-decimation procedure means that we do not depend on long-range connections between pixels.

\subsection{Aligning Segmentation Hierarchies}
\label{sec:align_seg}

In order to leverage multi-scale information, our approach combines segmentation hierarchies computed independently at multiple image resolutions. 
However, since subsampling an image removes details and smooths away boundaries, the resulting UCMs are misaligned, as illustrated in the second panel of Fig. 2. 
In this section, we propose an algorithm to align an arbitrary segmentation hierarchy to a target segmentation  
and, in Sect. 5, we show its effectiveness for multi-scale segmentation.



\begin{figure}[t]
\begin{center}
   \includegraphics[width=0.24\linewidth]{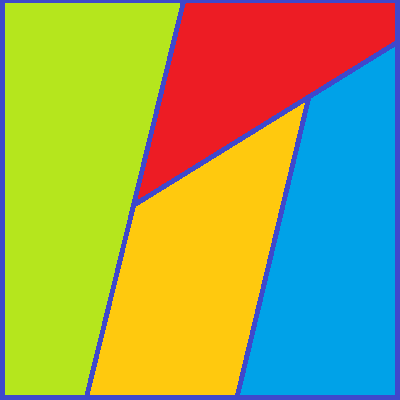}
   \includegraphics[width=0.24\linewidth]{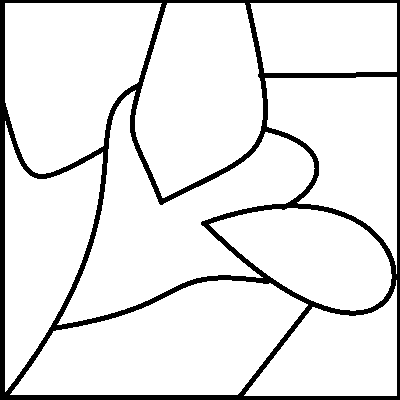}
   \includegraphics[width=0.24\linewidth]{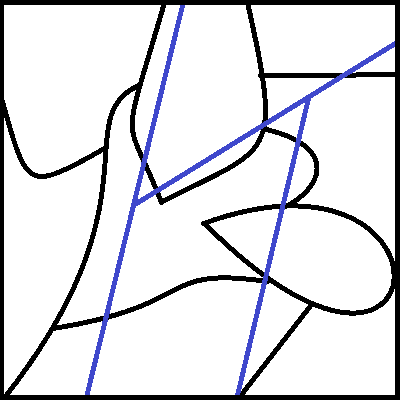}
   \includegraphics[width=0.24\linewidth]{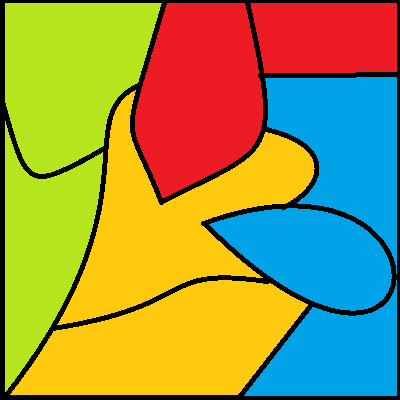}
\end{center}
   \caption{\textbf{Example of segmentation projection}. In order to ``snap'' the boundaries of a segmentation $\mathcal{R}$ (left) to those of a segmentation $\mathcal{S}$ (middle), since they do not align, we compute $\pi( \mathcal{R}, \mathcal{S})$ (right) by assigning to each segment in $\mathcal{S}$ its mode among the labels of $\mathcal{R}$.}
\label{fig:ucmT}
\end{figure}

The basic operation is to ``snap'' the boundaries of a segmentation $\mathcal{R} = \{ R_i\}_i$ to a segmentation $\mathcal{S} = \{ S_j\}_j$, as illustrated in Fig. \ref{fig:ucmT}. 
For this purpose, we define $\mathcal{L}(S_j)$, the new label of a region $S_j \in \mathcal{S}$, as the majority label of its pixels in $\mathcal{R}$:
\begin{equation}
\mathcal{L}(S_j) = \argmax_i \frac{| S_j \cap R_i | }{|S_j|}
\end{equation}
We call the segmentation defined by this new labeling of all the regions of $\mathcal{S}$ the \emph{projection} of $\mathcal{R}$ onto $\mathcal{S}$ and denote it by $\pi(\mathcal{R}, \mathcal{S})$.


In order to project an UCM onto a target segmentation $\mathcal{S}$, which we denote $\pi(\textrm{UCM}, \mathcal{S})$, we project in turn each of the levels of the hierarchy onto $\mathcal{S}$.
Note that, since all the levels are projected onto the same segmentation, the family of projections is by construction a hierarchy of segmentations.
This procedure is summarized in pseudo-code in Algorithm \ref{alg:ucm_S}.


\begin{algorithm}[h]
\caption{UCM Rescaling and Alignment}
\begin{algorithmic}[1]
\Require An UCM with a set of levels $\left[ t_1,...,t_K \right]$
\Require A target segmentation $\mathcal{S}^{*}$
\State $\textrm{UCM}_\pi \gets 0$
\For{$t = \left[ t_1,...,t_K \right]$}
\State $\mathcal{S} \gets {\tt sampleHierarchy}(\textrm{UCM},t)$
\State $\mathcal{S} \gets {\tt rescaleSegmentation}(\mathcal{S},\mathcal{S}^{*})$
\State $\mathcal{S} \gets \pi(\mathcal{S},\mathcal{S}^{*})$
\State $contours \gets {\tt extractBoundary}(\mathcal{S})$  
\State $\textrm{UCM}_\pi \gets {\tt max}(\textrm{UCM}_\pi, t*contours)$ 
\EndFor
\State \Return $\textrm{UCM}_\pi$
\end{algorithmic}
\label{alg:ucm_S}
\end{algorithm}

Observe that the routines ${\tt sampleHierarchy}$ and ${\tt extractBoundary}$ can be computed efficiently because they involve only thresholding operations and connected components labeling. 
The complexity is thus dominated by rescaleSegmentation in Step 4, a nearest neighbor interpolation, and the projection in Step 5, which are computed $K$ times.

\subsection{Multiscale Hierarchical Segmentation}
\label{sect:multi}

\paragraph*{\textbf{Single-scale segmentation}} 
We consider as input the following local contour cues: (1) brightness, color and texture differences in half-disks of three sizes \cite{Martin-etc:PAMI}, (2) sparse coding on patches \cite{renNIPS12}, and (3) structured forest contours \cite{Dollar:ICCV13}.
We globalize the contour cues independently using our fast eigenvector gradients of Sect. \ref{sec:fast_eigen}, combine global and local cues linearly, and construct an UCM based on the mean contour strength.  
We tried learning weights using gradient ascent on the F-measure on the
training set~\cite{Arbelaez2011}, but evaluating the final hierarchies rather than open contours.
We observed that this objective favors the quality of contours at the expense of regions and obtained better overall results by optimizing the Segmentation Covering metric~\cite{Arbelaez2011}.

\paragraph*{\textbf{Hierarchy Alignment}}
We construct a multiresolution pyramid with $N$ scales by subsampling / supersampling the original image and applying our single-scale segmenter. 
In order to preserve thin structures and details, we declare as set of possible boundary locations the finest superpixels in the highest-resolution. 
Then, applying recursively Algorithm \ref{alg:ucm_S}, we project each coarser UCM onto the next finer scale until aligning it to the highest resolution superpixels.

\paragraph*{\textbf{Multiscale Hierarchy}}
After alignment, we have a fixed set of boundary locations, and $N$ strengths for each of them, coming from the different scales. We formulate this problem as binary boundary classification and train a classifier that combines these $N$ features into a single probability of boundary estimation.
We experimented with several learning strategies for combining UCM strengths: (a) Uniform weights transformed into probabilities with Platt's method. 
(b) SVMs and logistic regression, with both linear and additive kernels. (c) Random Forests. (d) The same algorithm as for single-scale.
We found the results with all learning methods surprisingly similar, in agreement with the observation reported by \cite{Martin-etc:PAMI}.
This particular learning problem, with only a handful of dimensions and millions of data points, is relatively easy and performance is mainly driven by our already high-performing and well calibrated features. 
We therefore use the simplest option (a). 

\section{Experiments on the BSDS500}
\label{sect:bsds}
We conduct extensive experiments on the BSDS500 \cite{wwwBSDS}, using the standard evaluation metrics
and following the best practice rules of that dataset.
We also report results with a recent evaluation metric $F_\mathit{op}$~\cite{Pont-Tuset2015c,jordi:cvpr2013},
Precision-Recall for objects and parts, using the publicly-available code.

\paragraph*{\textbf{Single-scale Segmentation}}
Table \ref{tab:bsds_benchmarks}-top shows the performance of our single-scale segmenter for different types of input contours on the validation set of the BSDS500. 
We obtain high-quality hierarchies for all the cues considered, showing the generality of our approach.
Furthermore, when using them jointly (row 'Comb.' in top panel), our segmenter outperforms the versions with individual cues, suggesting its ability to leverage diversified inputs.
In terms of efficiency, our fast normalized cuts algorithm provides an average 20$\times$ speed-up over \cite{Arbelaez2011}, starting from the same local cues, with no significant loss in accuracy and with a low memory footprint. 

\paragraph*{\textbf{Multiscale Segmentation}}
Table \ref{tab:bsds_benchmarks}-bottom evaluates our full approach in the same experimental conditions as the upper panel. 
We observe a consistent improvement in performance in all the metrics for all the inputs, which validates our architecture for multiscale segmentation.  
We experimented with the range of scales and found $N=\{0.5, 1, 2\}$ adequate for our purposes. A finer sampling or a wider range of scales did not provide noticeable improvements.  
We tested also two degraded versions of our system (not shown in the table). For the first one, we resized contours to the original image resolution, created UCMs and combined them with the same method as our final system. For the second one, we transformed per-scale UCMs to the original resolution, but omitted the strength transfer to the finest superpixels before combining them. The first ablated version produces interpolation artifacts and smooths away details, while the second one suffers from misalignment. Both fail to improve performance over the single-scale result, which provides additional empirical support for our multiscale approach. We also observed a small degradation in performance when forcing the input contour detector to use only the original image resolution, which indicates the advantages of considering multiscale information at all stages of processing.

Since there are no drastic changes in our results when taking as input the different individual cues or their combination, in the sequel we use the version with structured forests for efficiency reasons, which we denote \textit{MCG-UCM-Our}.

\begin{figure*}
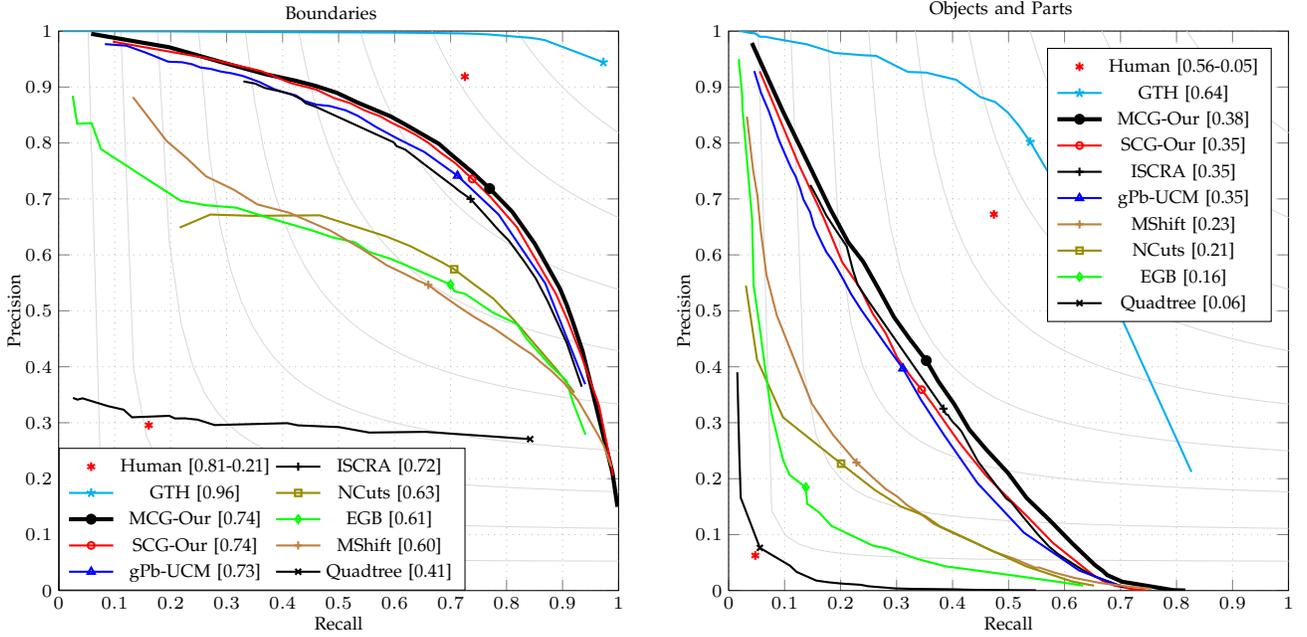

\centering
\scalebox{0.98}{%
\begin{minipage}[b]{0.46\linewidth}
\centering
\scriptsize \hspace{6mm}Boundaries
\begin{tikzpicture}[/pgfplots/width=1.1\linewidth, /pgfplots/height=1.1\linewidth, /pgfplots/legend pos=south west]
    \begin{axis}[ymin=0,ymax=1,xmin=0,xmax=1,enlargelimits=false,
        xlabel=Recall,
        ylabel=Precision,
        font=\scriptsize, grid=major,
        grid style=dotted,
        axis equal image=true,
        legend columns=5,
        transpose legend,
        legend style={at={(0,0)},
        anchor=south west},
        xlabel shift={-2pt},
        ylabel shift={-3pt},
        xtick={0,0.1,0.2,0.3,0.4,0.5,0.6,0.7,0.8,0.9,1},
        ytick={0,0.1,0.2,0.3,0.4,0.5,0.6,0.7,0.8,0.9,1}]
    \addplot[white!85!black,line width=0.2pt,domain=(0.1/(2-0.1)):1,samples=40,forget plot]{(0.1*x)/(2*x-0.1)};
    \addplot[white!85!black,line width=0.2pt,domain=(0.2/(2-0.2)):1,samples=40,forget plot]{(0.2*x)/(2*x-0.2)};
    \addplot[white!85!black,line width=0.2pt,domain=(0.3/(2-0.3)):1,samples=40,forget plot]{(0.3*x)/(2*x-0.3)};
    \addplot[white!85!black,line width=0.2pt,domain=(0.4/(2-0.4)):1,samples=40,forget plot]{(0.4*x)/(2*x-0.4)};
    \addplot[white!85!black,line width=0.2pt,domain=(0.5/(2-0.5)):1,samples=40,forget plot]{(0.5*x)/(2*x-0.5)};
    \addplot[white!85!black,line width=0.2pt,domain=(0.6/(2-0.6)):1,samples=40,forget plot]{(0.6*x)/(2*x-0.6)};
    \addplot[white!85!black,line width=0.2pt,domain=(0.7/(2-0.7)):1,samples=40,forget plot]{(0.7*x)/(2*x-0.7)};
    \addplot[white!85!black,line width=0.2pt,domain=(0.8/(2-0.8)):1,samples=40,forget plot]{(0.8*x)/(2*x-0.8)};
    \addplot[white!85!black,line width=0.2pt,domain=(0.9/(2-0.9)):1,samples=40,forget plot]{(0.9*x)/(2*x-0.9)};

   \addplot+[only marks,red,mark=asterisk,mark size=1.7,thick,forget plot] table[x=Recall,y=Precision] {data/pr_curves/fb_human_diff.txt};
    \addplot+[only marks,red,mark=asterisk,mark size=1.7,thick] table[x=Recall,y=Precision] {data/pr_curves/fb_human.txt};
    \addlegendentry{Human [%
\mbox{\input{data/pr_curves/fb_human_ods_f.txt}\hspace{-2.5pt}}%
-%
\mbox{\input{data/pr_curves/fb_human_diff_ods_f.txt}\hspace{-2.5pt}}%
]}

    \addplot+[cyan,solid,mark=none, thick,forget plot] table[x=Recall,y=Precision] {data/segm_bsds/test_fb_gt_sp150_ucm2.txt};
    \addplot+[cyan,solid,mark=star,mark size=1.8, thick] table[x=Recall,y=Precision] {data/segm_bsds/test_fb_gt_sp150_ucm2_ods.txt};
    \addlegendentry{GTH [%
\mbox{\input{data/segm_bsds/test_fb_gt_sp150_ucm2_ods_f.txt}\hspace{-2.5pt}}%
]}

    \addplot+[black,solid,mark=none,ultra thick,forget plot] table[x=Recall,y=Precision] {data/segm_bsds/test_fb_MCG-ucm.txt};
    \addplot+[black,solid,mark=o, mark size=1.3,ultra thick] table[x=Recall,y=Precision] {data/segm_bsds/test_fb_MCG-ucm_ods.txt};
    \addlegendentry{MCG-Our [%
\mbox{\input{data/segm_bsds/test_fb_MCG-ucm_ods_f.txt}\hspace{-2.5pt}}%
]}
   
    \addplot+[red,solid,mark=none, thick,forget plot] table[x=Recall,y=Precision] {data/segm_bsds/test_fb_SCG-ucm.txt};
    \addplot+[red,solid,mark=o, mark size=1.3, thick] table[x=Recall,y=Precision] {data/segm_bsds/test_fb_SCG-ucm_ods.txt};
    \addlegendentry{SCG-Our [%
\mbox{\input{data/segm_bsds/test_fb_SCG-ucm_ods_f.txt}\hspace{-2.5pt}}%
]}
    
    \addplot+[blue,solid,mark=none, thick,forget plot] table[x=Recall,y=Precision] {data/segm_bsds/test_fb_UCM.txt};
    \addplot+[blue,solid,mark=triangle, mark size=1.6, thick] table[x=Recall,y=Precision] {data/segm_bsds/test_fb_UCM_ods.txt};
    \addlegendentry{gPb-UCM [%
\mbox{\input{data/segm_bsds/test_fb_UCM_ods_f.txt}\hspace{-2.5pt}}%
]}

    \addplot+[black,solid,mark=none, thick,forget plot] table[x=Recall,y=Precision] {data/segm_bsds/test_fb_iscra.txt};
    \addplot+[black,solid,mark=+, mark size=1.6, thick] table[x=Recall,y=Precision] {data/segm_bsds/test_fb_iscra_ods.txt};
    \addlegendentry{ISCRA [%
\mbox{\input{data/segm_bsds/test_fb_iscra_ods_f.txt}\hspace{-2.5pt}}%
]}

    \addplot+[olive,solid,mark=none, thick,forget plot] table[x=Recall,y=Precision] {data/pr_curves/fb_NCut.txt};
    \addplot+[olive,solid,mark=square, mark size=1.25, thick] table[x=Recall,y=Precision] {data/pr_curves/fb_NCut_ods.txt};
    \addlegendentry{NCuts [%
\mbox{\input{data/pr_curves/fb_NCut_ods_f.txt}\hspace{-2.5pt}}%
]}
    
       \addplot+[green,solid,mark=none, thick,forget plot] table[x=Recall,y=Precision] {data/pr_curves/fb_FelzHutt.txt};
    \addplot+[green,solid,mark=diamond, mark size=1.5, thick] table[x=Recall,y=Precision] {data/pr_curves/fb_FelzHutt_ods.txt};
    \addlegendentry{EGB [%
\mbox{\input{data/pr_curves/fb_FelzHutt_ods_f.txt}\hspace{-2.5pt}}%
]} 
    
    \addplot+[brown,solid,mark=none, thick,forget plot] table[x=Recall,y=Precision] {data/pr_curves/fb_MeanShift.txt};
    \addplot+[brown,solid,mark=+, mark size=1.6, thick] table[x=Recall,y=Precision] {data/pr_curves/fb_MeanShift_ods.txt};
    \addlegendentry{MShift [%
\mbox{\input{data/pr_curves/fb_MeanShift_ods_f.txt}\hspace{-2.5pt}}%
]}




    \addplot+[black,solid,mark=none,solid,thick,forget plot] table[x=Recall,y=Precision] {data/pr_curves/fb_QuadTree.txt};
    \addplot+[black,solid,mark=x,mark size=1.6,solid,thick] table[x=Recall,y=Precision] {data/pr_curves/fb_QuadTree_ods.txt};
    \addlegendentry{Quadtree [%
\mbox{\input{data/pr_curves/fb_QuadTree_ods_f.txt}\hspace{-2.5pt}}%
]}
  \end{axis}
\end{tikzpicture}
\end{minipage}
\hspace{5mm}
\begin{minipage}[b]{0.46\linewidth}
\centering
\scriptsize \hspace{6mm}Objects and Parts
\begin{tikzpicture}[/pgfplots/width=1.1\linewidth, /pgfplots/height=1.1\linewidth, /pgfplots/legend pos=south west]
    \begin{axis}[ymin=0,ymax=1,xmin=0,xmax=1,enlargelimits=false,
        xlabel=Recall,
        ylabel=Precision,
        font=\scriptsize, grid=major,
        legend pos=north east,
        grid style=dotted,
        axis equal image=true,
        xlabel shift={-2pt},
        ylabel shift={-3pt},
        xtick={0,0.1,0.2,0.3,0.4,0.5,0.6,0.7,0.8,0.9,1},
        ytick={0,0.1,0.2,0.3,0.4,0.5,0.6,0.7,0.8,0.9,1}]
]
    \addplot[white!85!black,line width=0.2pt,domain=(0.1/(2-0.1)):1,samples=40,forget plot]{(0.1*x)/(2*x-0.1)};
    \addplot[white!85!black,line width=0.2pt,domain=(0.2/(2-0.2)):1,samples=40,forget plot]{(0.2*x)/(2*x-0.2)};
    \addplot[white!85!black,line width=0.2pt,domain=(0.3/(2-0.3)):1,samples=40,forget plot]{(0.3*x)/(2*x-0.3)};
    \addplot[white!85!black,line width=0.2pt,domain=(0.4/(2-0.4)):1,samples=40,forget plot]{(0.4*x)/(2*x-0.4)};
    \addplot[white!85!black,line width=0.2pt,domain=(0.5/(2-0.5)):1,samples=40,forget plot]{(0.5*x)/(2*x-0.5)};
    \addplot[white!85!black,line width=0.2pt,domain=(0.6/(2-0.6)):1,samples=40,forget plot]{(0.6*x)/(2*x-0.6)};
    \addplot[white!85!black,line width=0.2pt,domain=(0.7/(2-0.7)):1,samples=40,forget plot]{(0.7*x)/(2*x-0.7)};
    \addplot[white!85!black,line width=0.2pt,domain=(0.8/(2-0.8)):1,samples=40,forget plot]{(0.8*x)/(2*x-0.8)};
    \addplot[white!85!black,line width=0.2pt,domain=(0.9/(2-0.9)):1,samples=40,forget plot]{(0.9*x)/(2*x-0.9)};

       \addplot+[only marks,red,mark=asterisk,mark size=1.7,thick,forget plot] table[x=Recall,y=Precision] {data/pr_curves/pro_human_diff.txt};
    \addplot+[only marks,red,mark=asterisk,mark size=1.7,thick] table[x=Recall,y=Precision] {data/pr_curves/pro_human.txt};
    \addlegendentry{Human [%
\mbox{\input{data/pr_curves/pro_human_ods_f.txt}\hspace{-2.5pt}}%
-%
\mbox{\input{data/pr_curves/pro_human_diff_ods_f.txt}\hspace{-2.5pt}}%
]}

    \addplot+[cyan,solid,mark=none, thick,forget plot] table[x=Recall,y=Precision] {data/segm_bsds/test_opf_gt_sp150_ucm2.txt};
    \addplot+[cyan,solid,mark=star,mark size=1.8, thick] table[x=Recall,y=Precision] {data/segm_bsds/test_opf_gt_sp150_ucm2_ods.txt};
    \addlegendentry{GTH [%
\mbox{\input{data/segm_bsds/test_opf_gt_sp150_ucm2_ods_f.txt}\hspace{-2.5pt}}%
]}

    \addplot+[black,solid,mark=none, ultra thick,forget plot] table[x=Recall,y=Precision] {data/segm_bsds/test_fop_MCG-ucm.txt};
    \addplot+[black,solid,mark=o, mark size=1.3, ultra thick] table[x=Recall,y=Precision] {data/segm_bsds/test_fop_MCG-ucm_ods.txt};
    \label{bsds:mcg}
    \addlegendentry{MCG-Our [%
\mbox{\input{data/segm_bsds/test_fop_MCG-ucm_ods_f.txt}\hspace{-2.5pt}}%
]}

    \addplot+[red,solid,mark=none, thick,forget plot] table[x=Recall,y=Precision] {data/segm_bsds/test_fop_SCG-ucm.txt};
    \addplot+[red,solid,mark=o, mark size=1.3, thick] table[x=Recall,y=Precision] {data/segm_bsds/test_fop_SCG-ucm_ods.txt};
    \label{bsds:scg}
    \addlegendentry{SCG-Our [%
\mbox{\input{data/segm_bsds/test_fop_SCG-ucm_ods_f.txt}\hspace{-2.5pt}}%
]}
    
    \addplot+[black,solid,mark=none, thick,forget plot] table[x=Recall,y=Precision] {data/segm_bsds/test_opf_iscra.txt};
    \addplot+[black,solid,mark=+, mark size=1.6, thick] table[x=Recall,y=Precision] {data/segm_bsds/test_opf_iscra_ods.txt};
    \label{bsds:iscra}
    \addlegendentry{ISCRA [%
\mbox{\input{data/segm_bsds/test_opf_iscra_ods_f.txt}\hspace{-2.5pt}}%
]}

    \addplot+[blue,solid,mark=none, thick,forget plot] table[x=Recall,y=Precision] {data/segm_bsds/test_opf_UCM.txt};
    \addplot+[blue,solid,mark=triangle, mark size=1.6, thick] table[x=Recall,y=Precision] {data/segm_bsds/test_opf_UCM_ods.txt};
    \label{bsds:gpb}
    \addlegendentry{gPb-UCM [%
\mbox{\input{data/segm_bsds/test_opf_UCM_ods_f.txt}\hspace{-2.5pt}}%
]}

    \addplot+[brown,solid,mark=none, thick,forget plot] table[x=Recall,y=Precision] {data/pr_curves/pro_MeanShift.txt};
    \addplot+[brown,solid,mark=+,mark size=1.6, thick] table[x=Recall,y=Precision] {data/pr_curves/pro_MeanShift_ods.txt};
    \addlegendentry{MShift [%
\mbox{\input{data/pr_curves/pro_MeanShift_ods_f.txt}\hspace{-2.5pt}}%
]}


    \addplot+[olive,solid,mark=none, thick,forget plot] table[x=Recall,y=Precision] {data/pr_curves/pro_NCut.txt};
    \addplot+[olive,solid,mark=square, mark size=1.25, thick] table[x=Recall,y=Precision] {data/pr_curves/pro_NCut_ods.txt};
    \addlegendentry{NCuts [%
\mbox{\input{data/pr_curves/pro_NCut_ods_f.txt}\hspace{-2.5pt}}%
]}


    \addplot+[green,solid,mark=none, thick,forget plot] table[x=Recall,y=Precision] {data/pr_curves/pro_FelzHutt.txt};
    \addplot+[green,solid,mark=diamond, mark size=1.5, thick] table[x=Recall,y=Precision] {data/pr_curves/pro_FelzHutt_ods.txt};
    \addlegendentry{EGB [%
\mbox{\input{data/pr_curves/pro_FelzHutt_ods_f.txt}\hspace{-2.5pt}}%
]}

    \addplot+[black,solid,mark=none,solid,thick,forget plot] table[x=Recall,y=Precision] {data/pr_curves/pro_QuadTree.txt};
    \addplot+[black,solid,mark=x,mark size=1.6,solid,thick] table[x=Recall,y=Precision] {data/pr_curves/pro_QuadTree_ods.txt};
    \addlegendentry{Quadtree [%
\mbox{\input{data/pr_curves/pro_QuadTree_ods_f.txt}\hspace{-2.5pt}}%
]}

    \end{axis}
\end{tikzpicture}
\end{minipage}\hspace{3mm}}
\caption{\textbf{BSDS500 test set.} Precision-Recall curves for boundaries~\cite{wwwBSDS} (left) and for objects and parts~\cite{Pont-Tuset2015c} (right). The marker on each curve is placed on the Optimal Dataset Scale (ODS), and its F measure is presented in brackets in the legend.
The isolated red asterisks refer to the human performance assessed on the same image (one human partition against the rest of human annotations on the same image) and on a different image (one human partition against the human partitions of a different, randomly selected, image).}
\label{fig:pr_curves}
\end{figure*}

\begin{table}
\begin{center}
\scalebox{0.65}{%
\begin{tabular}{cl||cc||cc|cc|cc|cc}
\cmidrule[\heavyrulewidth](l{-2pt}){3-12}
&\multicolumn{1}{c}{} &   \multicolumn{2}{c||}{Boundary}&  \multicolumn{8}{c}{Region} \\
\cmidrule(l{-2pt}){3-12}
&\multicolumn{1}{c}{} &   \multicolumn{2}{c||}{$F_b$}&  \multicolumn{2}{c|}{$F_{op}$} & \multicolumn{2}{c|}{SC}&  \multicolumn{2}{c|}{PRI} &  \multicolumn{2}{c}{VI} \\
&\multicolumn{1}{c}{Input} &   ODS & OIS & ODS & OIS & ODS & OIS & ODS & OIS & ODS & OIS\\
\midrule[\heavyrulewidth]
\multirow{4}{*}{\rotatebox{90}{\footnotesize Single-Scale}\vspace{-0.02in}} &Pb \cite{Martin-etc:PAMI} & 0.702 & 0.733 & 0.334 & 0.370 & 0.577 & 0.636 & 0.801 & 0.847 & 1.692 & 1.490\\[1mm]
&SC \cite{renNIPS12}& 0.697 & 0.725 & 0.264 & 0.306 & 0.540 & 0.607 & 0.777 & 0.835 & 1.824 & 1.659\\[1mm]
&SF \cite{Dollar:ICCV13} & 0.719 & 0.737 & 0.338 & 0.399 & 0.582 & 0.651 & 0.803 & 0.851 & 1.608 & 1.432\\[1mm]
&Comb. & 0.719 & 0.750 & 0.358 & 0.403 & 0.602 & 0.655 & 0.809 & 0.855 & 1.619 & 1.405\\
\midrule
\multirow{4}{*}{\rotatebox{90}{\small Multiscale}\vspace{-0.02in}} &Pb \cite{Martin-etc:PAMI} & 0.713 & 0.745 & 0.350 & 0.389 & 0.598 & 0.656 & 0.807 & 0.856 & 1.601 & 1.418\\[1mm]
&SC \cite{renNIPS12} & 0.705 & 0.734 & 0.331 & 0.384 & 0.579 & 0.647 & 0.799 & 0.851 & 1.637 & 1.460\\[1mm]
&SF \cite{Dollar:ICCV13} & 0.725 & 0.744 & 0.370 & \textbf{0.420} & 0.600 & 0.660 & 0.810 & 0.854 & 1.557 & 1.390\\[1mm]
&Comb.  & \textbf{0.725} & \textbf{0.757} & \textbf{0.371} & 0.408 & \textbf{0.611} & \textbf{0.670} & \textbf{0.813} & \textbf{0.862} & \textbf{1.548} & \textbf{1.367}\\
\bottomrule
\end{tabular}}
\vspace{1mm}
\end{center}
   \caption{\textbf{BSDS500 val set.} Control experiments for single-scale (top) and multiscale (bottom) hierarchical segmentation with different input contour detectors}
   \label{tab:bsds_benchmarks}
\end{table}

\paragraph*{\textbf{Comparison with state-of-the-art.}}
Figure~\ref{fig:pr_curves} compares our multiscale hierarchical segmenter MCG~(\ref{bsds:mcg}) and our
single-scale hierarchical segmenter SCG~(\ref{bsds:scg}) on the BSDS500 test set against all the methods for
which there is publicly available code. 
We also compare to the recent ISCRA~\cite{Ren_2013_CVPR} hierarchies~(\ref{bsds:iscra}), provided precomputed by the authors.
We obtain consistently the best results to date on BSDS500 for all operating regimes, both in terms of boundary and region quality.

Note that the range of object scales in the BSDS500 is limited, which translates into modest absolute gains
from MCG~(\ref{bsds:mcg}) with respect to SCG~(\ref{bsds:scg}) in terms of boundary evaluation (left-hand plot),
but more significant improvements in terms of objects and parts (right-hand plot). 
We will also observe more substantial improvements with respect to gPb-UCM~(\ref{bsds:gpb}) when we move to PASCAL, SBD, and COCO in Section~\ref{sect:experim_prop} (e.g.\ see Fig. \ref{fig:J_i}).

\paragraph*{\textbf{Ground-Truth Hierarchy}}
In order to gain further insights, 
we transfer the strength of each ground-truth segmentation to our highest-resolution superpixels $\mathcal{S}^{N*}$ and construct a combined hierarchy.
This approximation to the semantic hierarchy, Ground-Truth Hierarchy (GTH) in Fig.~\ref{fig:pr_curves}, is an upper-bound for our approach as both share the same boundary locations and the only difference is their strength. 
Since the strength of GTH is proportional to the number of subjects marking it, it provides naturally the correct semantic ordering, where outer object boundaries are stronger than internal parts. 

Recently, Maire \etal \cite{Maire-etal:BMVC13} developed an annotation tool where the user encodes explicitly the ``perceptual strength'' of each contour. 
Our approach provides an alternative where the semantic hierarchy is reconstructed by sampling flat annotations from multiple subjects.

\section{Object Proposal Generation}
\label{sec:obj_prop}
The image segmentation algorithm presented in the previous sections builds on low-level features,
so its regions are unlikely to represent accurately complete objects with heterogeneous parts.
In this context, object proposal techniques create a set of hypotheses, possibly overlapping,
which are more likely to represent full object instances.

Our approach to object proposal generation is to combinatorially look for sets of regions from our segmentation hierarchies 
that merged together are likely to represent complete objects.
In this section, we first describe the efficient computation of certain region descriptors on a segmentation tree.
Then, we describe how we use these techniques to efficiently explore the sets of merged regions from the hierarchy.
Finally, we explain how we train the parameters of our algorithm for object proposals and
how we rank the candidates by their probability of representing an object.

\paragraph*{\textbf{Fast computation of descriptors}}
Let us assume, for instance, we want to compute the area of all regions in the hierarchy.
Intuitively, working strictly on the merging-sequence partitions, we would need to scan all pixels
in all partitions.
On the other hand, working on the region tree allows us to scan the image only once to compute the area of the leaves,
and then propagate the area to all parents as the addition of the areas of their children.

As a drawback, the algorithms become intricate in terms of coding and necessary data structures.
Take, for instance, the computation of the neighbors of a certain region,
which is trivial via scanning the partition on the merging sequence (look for region labels in the adjacent boundary pixels),
but need tailored data structures and algorithms in the region tree.

Formally, let us assume the image has $p$ pixels, and we build a hierarchy based on $s$ superpixels
(leaves of the tree), and $m$ mergings (different UCM strength values).
The cost of computing the area on all regions using the merging-sequence partitions
will be the cost of scanning all pixels in these partitions, thus $p\!\cdot\!(m\!+\!1)$.
In contrast, the cost using the region tree will involve scanning the image once, and then
propagating the area values, so $p\!+\!m$, which is notably faster.

We built tailored algorithms and data structures to compute the
bounding box, perimeter, and neighbors of a region using the region tree representation. 

\begin{figure}[h]
\input{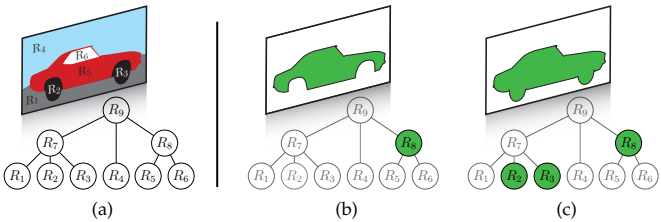}
\caption{\textbf{Object segmentation as combinatorial optimization}: Examples of objects (b), (c), formed by selecting regions from a hierarchy (a).}
\label{fig:reg_sel_examples}
\end{figure}

\paragraph*{\textbf{Combinatorial Grouping of Proposals}}
\label{sec:comb_cands}
We can cast object segmentation as \textit{selecting} some regions in the hierarchy, or in other words,
as a combinatorial optimization problem on the hierarchy.
To illustrate this principle, Figure~\ref{fig:reg_sel_examples}(a) shows the simplified representation
of the hierarchy in Figure~\ref{fig:hierarchy}.
Figure~\ref{fig:reg_sel_examples}(b) and (c) show two object proposals,
and their representation in the region hierarchy.

Since hierarchies are built taking only low-level features into account, and do not use semantic information,
objects will usually not be optimally represented using a single region in the hierarchy.
As an example, Figure~\ref{fig:reg_sel_examples}(c) shows the optimum representation of the car, consisting of three regions.

A sensible approach to create object proposals is therefore to explore the set of $n$-tuples of regions.
The main idea behind MCG is to explore this set efficiently, taking advantage of the region tree representation,
via the fast computation of region neighbors.

The whole set of tuples, however, is huge, and so it is not feasible to explore it exhaustively.
Our approach ranks the proposals using the height of their regions in the tree (UCM strength)
and explores the tree from the top, but only down to a certain threshold.
To speed the process up, we design a top-down algorithm to compute the region neighbors and thus only
compute them down to a certain depth in the tree.

To further improve the results, we not only consider the $n$-tuples from the resulting MCG-UCM-Our hierarchy, but also
the rest of hierarchies computed at different scales.
As we will show in the experiments, diversity significantly improves the proposal results.

\paragraph*{\textbf{Parameter Learning via Pareto Front Optimization}}
MCG takes a set of diverse hierarchies and computes the $n$-tuples up to a certain UCM strength. 
We can interpret the $n$-tuples from each hierarchy as a ranked list of $N_i$ proposals that are put
together to create the final set of $N_p$ proposals.

At training time, we would like to find, for different values of $N_p$, the number of proposals
from each ranked list $\overline{N}_i$ such that 
the joint pool of $N_p$ proposals has the best achievable quality.
We frame this learning problem as a Pareto front optimization~\cite{Everingham2006,Ehrgott2005} with two conflicting objective functions:
number of proposals and achievable quality.
At test time, we select a working point on the Pareto front, represented by the $\left\{\overline{N}_i\right\}$ values,
based either on the number of proposals $N_p$ we can handle or on the minimum achievable quality our application needs,
and we combine the $\overline{N}_i$ top proposals from each hierarchy list.

Formally, assuming $R$ ranked lists $L_i$, an exhaustive learning algorithm would consider
all possible values of the $R$-tuple $\left\{N_1,\dots,N_R\right\}$, where $N_i\in\left\{0,\dots,|L_i|\right\}$;
adding up to $\prod_1^R \left|L_i\right|$ parameterizations to try,
which is intractable in our setting.

\begin{figure}[h!]
   \includegraphics[width=\linewidth]{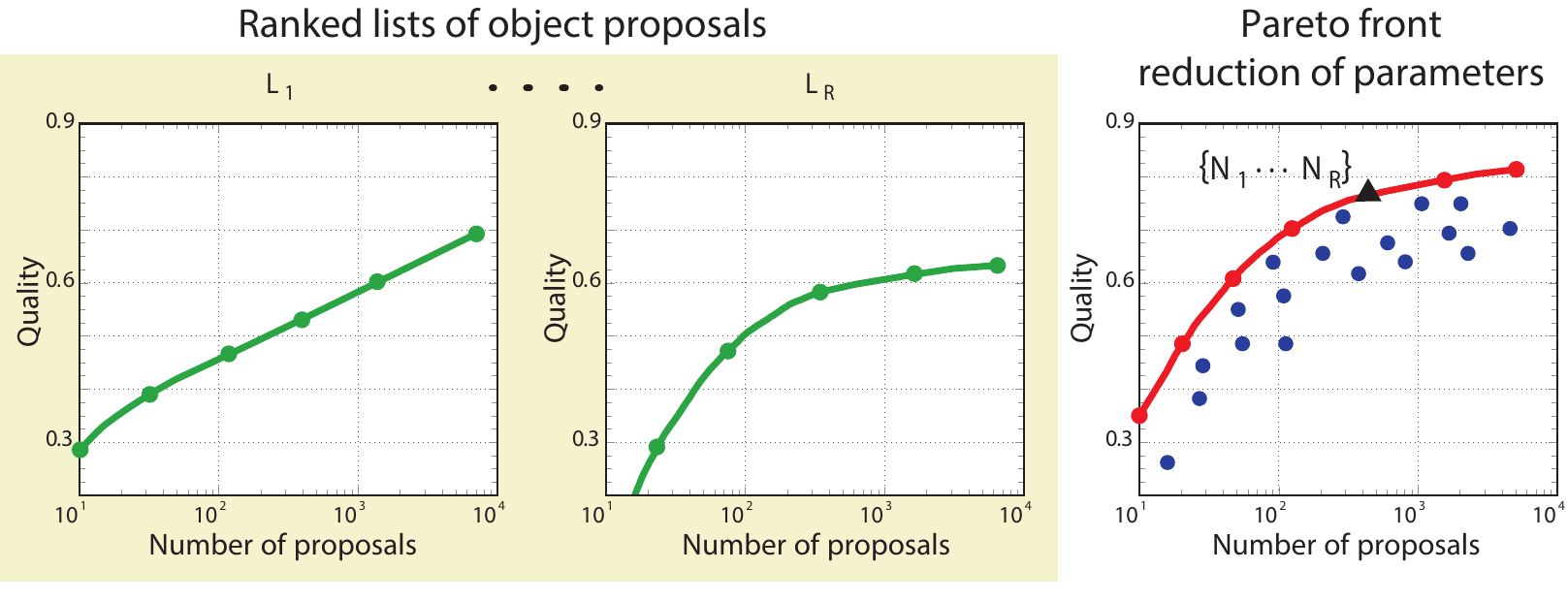}
   \caption{\textbf{Pareto front learning}:
   Training the combinatorial generation of proposals using the Pareto front}
   \label{fig:pareto}
\end{figure}

Figure~\ref{fig:pareto} illustrates the learning process.
To reduce the dimensionality of the search space, we start by selecting two ranked lists $L_1$, $L_2$ (green curves)
and we sample the list at $S$ levels of number of proposals (green dots).
We then scan the full $S^2$ different parameterizations to combine the proposals from both (blue dots).
In other words, we analyze the sets of proposals created by combining the top $N_1$ from $L_1$ (green dots) and
the top $N_2$ from $L_2$.

The key step of the optimization consists in discarding those parameterizations whose quality point is not
in the Pareto front (red curve).
(i.e., those parameterizations that can be substituted by another with better quality
with the same number of proposals, or by one with the same quality with less proposals.)
We sample the Pareto front to $S$ points and we iterate the process until all the ranked lists are combined.

Each point in the final Pareto front corresponds to a particular parameterization $\left\{N_1,\dots,N_R\right\}$.
At train time, we choose a point on this curve, either at a given number of proposals $N_c$ or at the achievable
quality we are interested in (black triangle) and store the parameters $\left\{\overline{N}_1,\dots,\overline{N}_R\right\}$.
At test time, we combine the $\left\{\overline{N}_1,\dots,\overline{N}_R\right\}$ top proposals from each ranked list.
The number of sampled configurations using the proposed algorithm is $(R\!-\!1)S^2$, that is, we have reduced an
exponential problem ($S^R$) to a quadratic one.

\paragraph*{\textbf{Regressed Ranking of Proposals}}
\label{sec:ranking}
To further reduce the number of proposals, we train a regressor from low-level features, as in~\cite{Carreira2012b}.
Since the proposals are all formed by a set of regions from a reduced set of hierarchies,
we focus on features that can be computed efficiently in a bottom-up fashion, as explained previously.

We compute the following features:
\begin{asparaitem}
\item\textbf{Size and location}: Area and perimeter of the candidate; area, position, and aspect ratio of the bounding box; and the
area balance between the regions in the candidate.
\item\textbf{Shape}: Perimeter (and sum of contour strength) divided by the squared root of the area; and
area of the region divided by that of the bounding box.
\item\textbf{Contours}: Sum of contour strength at the boundaries, mean contour strength at the boundaries; minimum and maximum UCM threshold of appearance and disappearance of the regions forming the candidate.
\end{asparaitem}
We train a Random Forest using these features to regress the object overlap with the ground truth, and
diversify the ranking based on Maximum Marginal Relevance measures~\cite{Carreira2012b}.
We tune the random forest learning on half training set and validating on the other half.
For the final results, we train on the training set and evaluate our proposals on the validation set of PASCAL  2012.

\section{Experiments on PASCAL VOC, SBD, and COCO}
This section presents our large-scale empirical validation of the object proposal algorithm described in the previous section.
We perform experiments in three annotated databases, with a variety of measures that demonstrate the state-of-the-art 
performance of our algorithm.

\label{sect:experim_prop}
\paragraph*{\textbf{Datasets and Evaluation Measures}}
We conduct experiments in the following three annotated datasets:
the segmentation challenge of PASCAL 2012 Visual Object Classes (SegVOC12)~\cite{Everingham2012},
the Berkeley Semantic Boundaries Dataset (SBD)~\cite{Hariharan2011}, and the Microsoft Common Objects in Context (COCO)~\cite{Lin2014}.
They all consist of images with annotated objects of different categories.
Table~\ref{tab:dbs} summarizes the number of images and object instances in each database.

\begin{table}[h]
\centering
\begin{tabular}{l|ccc}
\cmidrule[\heavyrulewidth](l{-2pt}){2-4}
\multicolumn{1}{c}{}    & Number of  & Number of  & Number of  \\
\multicolumn{1}{c}{}    & Classes    &  Images    & Objects    \\
    \midrule
SegVOC12             &    20 &  \,\,\,2\,913  &   \,\,\,9\,847        \\
SBD                  &    20 &  12\,031  &  32\,172   \\
COCO                 &    80 &  123\,287\,\,\,  &    910\,983\,\,\,   \\
 \bottomrule
\end{tabular}
\vspace{1mm}
\caption{Sizes of the databases}
\label{tab:dbs}
\end{table}

Regarding the performance metrics, we measure the achievable quality with respect to the number of proposals, that is,
the quality we would have if an oracle selected the best proposal among the pool.
This aligns with the fact that object proposals are a preprocessing step for other algorithms that will represent and classify them.
We want, therefore, the achievable quality within the proposals to be as high as possible, while reducing the number of proposals to make the
final system as fast as possible.

As a measure of quality of a specific proposal with respect to an annotated object, we consider 
the Jaccard index $J$, also known as overlap or intersection over union; which is defined as the 
size of the intersection of the two pixel sets over the size of their union.

To compute the overall quality for the whole database, we first select the best
proposal for each annotated instance with respect to $J$.
The \textit{Jaccard index at instance level} ($J_i$) is then defined as the mean
best overlap for all the ground-truth instances in the database,
also known as Best Spatial Support score (BSS)~\cite{Malisiewicz2007} or Average Best Overlap (ABO)~\cite{Uijlings2013}.

Computing the mean of the best overlap on all objects, as done by $J_i$,
hides the distribution of quality among different objects.
As an example, $J_i\!=\!0.5$ can mean that the algorithm covers half the objects
perfectly and completely misses the other half, or
can also mean that all the objects are covered exactly at $J\!=\!0.5$. 
This information might be useful to decide which algorithm to use.
Computing a histogram of the best overlap would provide very rich information,
but then the resulting plot would be 3D (number of proposals, bins, and bin counts).
Alternatively, we propose to plot different percentiles of the histogram.

Interestingly, a certain percentile of the histogram of best overlaps consists in
computing the number of objects whose best overlap is above a certain
Jaccard threshold, which can be interpreted as the best achievable recall of the technique
over a certain threshold.
We compute the recall at three different thresholds: $J\!=\!0.5$, $J\!=\!0.7$, and $J\!=\!0.85$.

\paragraph*{\textbf{Learning Strategy Evaluation}}
We first estimate the loss in performance due to not sweeping all the possible
values of $\left\{N_1,\dots,N_R\right\}$ in the combination of proposal lists via the proposed greedy strategy.
To do so, we will compare this strategy with the full combination on a reduced problem to make
the latter feasible.
Specifically, we combine the 4 ranked lists coming from the singletons at all scales,
instead of the full 16 lists coming from singletons, pairs, triplets, and 4-tuples.
We also limit the search to 20\,000 proposals, further speeding the process up.

In this situation, the mean loss in achievable quality along the full curve of parameterization
is $J_i\!=\!0.0002$, with a maximum loss of $J_i\!=\!0.004$ ($0.74\%$).
In exchange, our proposed learning strategy on the full 16 ranked lists takes about 20 seconds to compute
on the training set of SegVOC12, while the singleton-limited full combination takes 4 days
(the full combination would take months).

\paragraph*{\textbf{Combinatorial Grouping}}
We now evaluate the Pareto front optimization strategy in the training set of SegVOC12.
As before, we extract the lists of proposals from the three scales and the multiscale hierarchy,
for singletons, pairs, triplets, and 4-tuples of regions, leading to 16 lists, ranked
by the minimum UCM strength of the regions forming each proposal.



\begin{figure*}
\resizebox{0.885\textwidth}{!}{\mbox{\begin{minipage}[b]{0.61\linewidth}
\centering
\scriptsize \hspace{5mm}Training\\[-0.1mm]
\begin{tikzpicture}[/pgfplots/width=\linewidth, /pgfplots/height=0.7\linewidth]
    \begin{axis}[ymin=0.35,ymax=0.9,xmin=20,xmax=1000000,enlargelimits=false,
        xlabel=Number of proposals,
        ylabel=Jaccard index at instance level ($J_i$),
        font=\scriptsize, grid=both,
        legend style={legend pos=south east,font=\scriptsize},
        grid style=dotted,
        axis equal image=false,
        ytick={0.2,0.3,0.4,0.5,0.6,0.7,0.8,0.9},
        minor ytick={0.2,0.225,...,0.9},
        major grid style={white!20!black},
        minor grid style={white!70!black},
        xlabel shift={-2pt},
        ylabel shift={-3pt},
        xmode=log]
          \addplot+[black,solid,mark=none, thick] table[x=ncands,y=jac_instance] {data/pareto/1_all_pascal2012_train2012.txt};
		  \label{fig:train:4tuples}
          \addplot+[magenta,solid,mark=none, thick] table[x=ncands,y=jac_instance] {data/pareto/2_up_to_triplets_pascal2012_train2012.txt};
          \label{fig:train:triplets}
     	  \addplot+[cyan,solid,mark=none, thick] table[x=ncands,y=jac_instance] {data/pareto/3_up_to_pairs_pascal2012_train2012.txt};
	  \label{fig:train:pairs}
     	  \addplot+[olive,solid,mark=none, thick] table[x=ncands,y=jac_instance] {data/pareto/4_all_singletons_pascal2012_train2012.txt};
	  \label{fig:train:singletons}
	       \addplot+[magenta,dashed,mark=none, thick] table[x=ncands,y=jac_instance] {data/pareto/all_equal_pascal2012_train2012.txt};

		  \addplot+[red,dashed,mark=none, thick] table[x=ncands,y=jac_instance] {data/pareto/5_mcg_singletons_pascal2012_train2012.txt};
		  \label{ours-multi-singletons}
		  \addplot+[green,dashed,mark=none, thick] table[x=ncands,y=jaco] {data/obj_cands/train2012_ucm.txt};
		  \addplot+[black,dashed,mark=none, thick] table[x=ncands,y=jaco] {data/obj_cands/train2012_quadtree.txt};

		  \addplot+[red,only marks,solid,mark=asterisk, thick] table[x=ncands,y=jac_instance] {data/obj_cands/train2012_multi_3sc_u_4r_pareto_selected_point.txt};
		  \addplot+[red,only marks,solid,mark=+, thick] table[x=ncands,y=jac_instance] {data/obj_cands/train2012_multi_3sc_u_4r_filt_point.txt};
		  \addplot+[black,solid,ultra thick,mark=none] table[x=ncands,y=jac_instance] {data/obj_cands/train2012_sf_mUCM_multi_3sc_u_4r_12k.txt};
\label{marker:jacc_train:regressed}
	 \end{axis}
   \end{tikzpicture}
\end{minipage}
\begin{minipage}[b]{0.61\linewidth}
\centering
\scriptsize \hspace{5mm}Validation\\[0.5mm]
\begin{tikzpicture}[/pgfplots/width=\linewidth, /pgfplots/height=0.7\linewidth]
    \begin{axis}[ymin=0.35,ymax=0.9,xmin=20,xmax=1000000,enlargelimits=false,
        xlabel=Number of proposals,
        ylabel=Jaccard index at instance level ($J_i$),
        font=\scriptsize, grid=both,
        legend style={legend pos=outer north east,font=\scriptsize},
        grid style=dotted,
        axis equal image=false,
        ytick={0.2,0.3,0.4,0.5,0.6,0.7,0.8,0.9},
        minor ytick={0.2,0.225,...,0.9},
        major grid style={white!20!black},
        minor grid style={white!70!black},
        xlabel shift={-2pt},
        ylabel shift={-3pt},
        xmode=log]
          \addplot+[black,solid,mark=none, thick] table[x=ncands,y=jac_instance] {data/pareto/1_all_pascal2012_val2012.txt};
\addlegendentry{Pareto up to 4-tuples}
\label{fig:train:4tuples}
          \addplot+[magenta,solid,mark=none, thick] table[x=ncands,y=jac_instance] {data/pareto/2_up_to_triplets_pascal2012_val2012.txt};
          \label{fig:train:triplets}
\addlegendentry{Pareto up to triplets}
     	  \addplot+[cyan,solid,mark=none, thick] table[x=ncands,y=jac_instance] {data/pareto/3_up_to_pairs_pascal2012_val2012.txt};
	  \label{fig:train:pairs}
\addlegendentry{Pareto up to pairs}
     	  \addplot+[olive,solid,mark=none, thick] table[x=ncands,y=jac_instance] {data/pareto/4_all_singletons_pascal2012_val2012.txt};
	  \label{fig:train:singletons}
\addlegendentry{Pareto only singletons}
		  \addplot+[red,dashed,mark=none, thick] table[x=ncands,y=jac_instance] {data/pareto/5_mcg_singletons_pascal2012_val2012.txt};
		  \label{ours-multi-singletons}
\addlegendentry{Raw Ours-multi singl.}
		  \addplot+[green,dashed,mark=none, thick] table[x=ncands,y=jaco] {data/obj_cands/val2012_ucm.txt};
\addlegendentry{Raw gPb-UCM singl.}
		  \addplot+[black,dashed,mark=none, thick] table[x=ncands,y=jaco] {data/obj_cands/val2012_quadtree.txt};
\addlegendentry{Raw Quadtree singl.}
      \addplot+[magenta,dashed,mark=none, thick] table[x=ncands,y=jac_instance] {data/pareto/all_equal_pascal2012_val2012.txt};
	  \label{fig:train:equal}
\addlegendentry{Equal distribution}

		  \addplot+[red,only marks,solid,mark=asterisk, thick] table[x=ncands,y=jac_instance] {data/obj_cands/val2012_pareto_point.txt};
\label{marker:jacc_train:sel_pareto}
\addlegendentry{Selected configuration}

		  \addplot+[red,only marks,solid,mark=+, thick] table[x=ncands,y=jac_instance] {data/obj_cands/val2012_mcg_point.txt};
\label{marker:jacc_train:filt_pareto}
\addlegendentry{Filtered candidates}

		  \addplot+[black,solid,ultra thick,mark=none] table[x=ncands,y=jac_instance] {data/obj_cands/val2012_mcg.txt};
\label{marker:jacc_train:regressed}
\addlegendentry{Regressed ranking}

	 \end{axis}
   \end{tikzpicture}
\end{minipage}}}
\caption{\textbf{Pareto front evaluation.} Achievable quality of our proposals for singletons, pairs, triplets, and 4-tuples; and the raw proposals from the hierarchies on PASCAL SegVOC12 training (left) and validation (right) sets.}
\label{fig:obj_cands_train}
\vspace{-2mm}
\end{figure*}
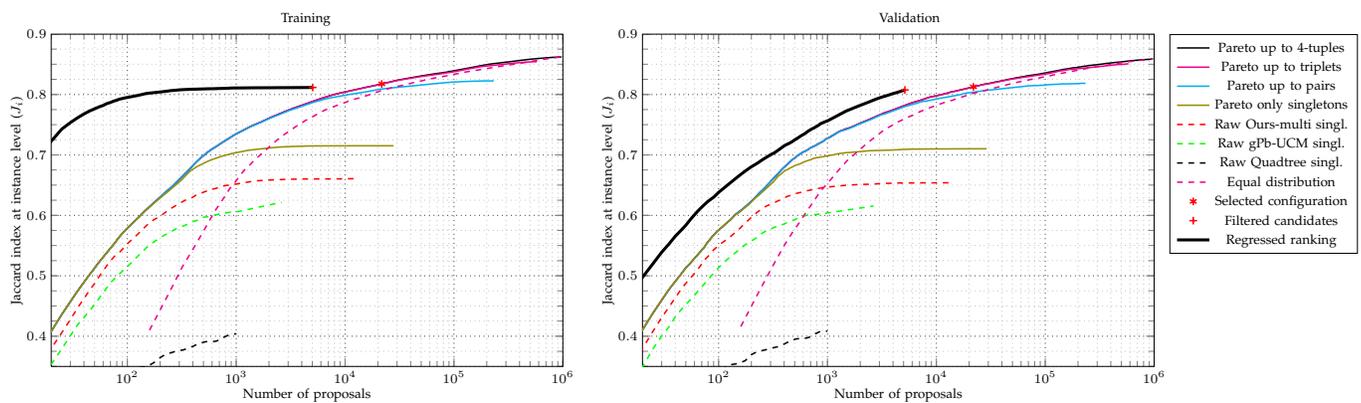

Figure~\ref{fig:obj_cands_train} shows the Pareto front evolution of $J_i$ with respect to the
number of proposals for up to 1, 2, 3, and 4 regions per proposal
(4, 8, 12, and 16 lists, respectively) at training and testing time on SegVOC12.
As baselines, we plot the raw singletons from MCG-UCM-Our, gPb-UCM, and Quadtree; as well as the uniform combination of scales.

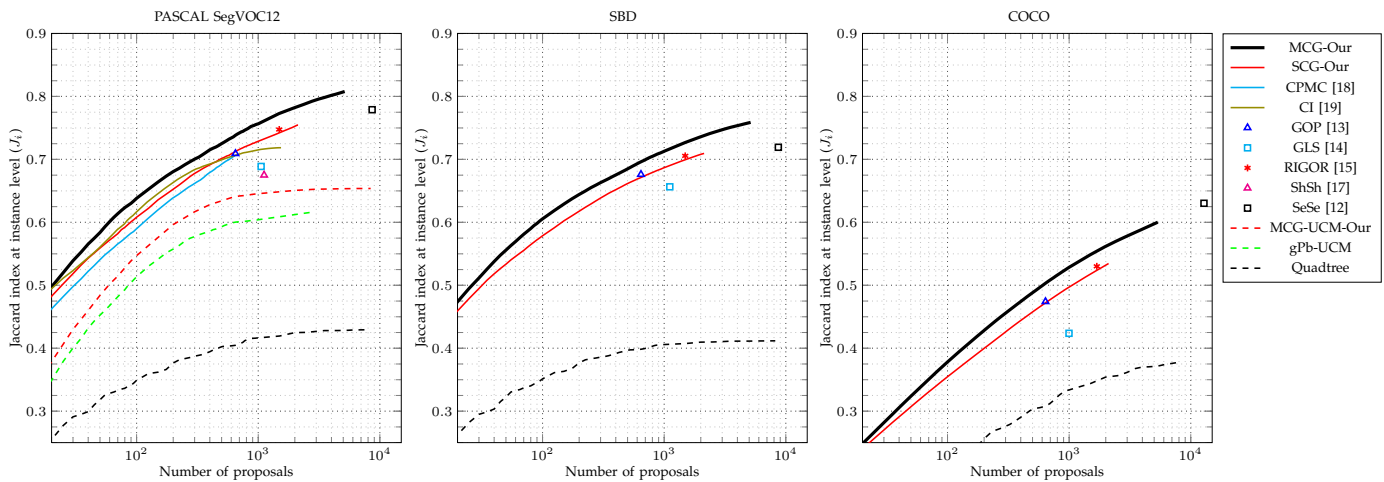
\begin{figure*}
\scalebox{0.75}{\hspace{-2mm}\begin{minipage}[b]{0.39\linewidth}
\centering
\scriptsize \hspace{5mm}PASCAL SegVOC12\\[-0.1mm]
\begin{tikzpicture}[/pgfplots/width=1.1\linewidth, /pgfplots/height=1.25\linewidth]
    \begin{axis}[ymin=0.25,ymax=0.9,xmin=20,xmax=15000,enlargelimits=false,
        xlabel=Number of proposals,
        ylabel=Jaccard index at instance level ($J_i$),
        font=\scriptsize, grid=both,
        grid style=dotted,
        axis equal image=false,
        ytick={0,0.1,...,1},
        minor ytick={0,0.025,...,1},
        major grid style={white!20!black},
        minor grid style={white!70!black},
        xlabel shift={-2pt},
        ylabel shift={-3pt},
        xmode=log]
		  \addplot+[black,solid,mark=none, ultra thick]                          table[x=ncands,y=jac_instance] {data/obj_cands/val2012_mcg.txt};
		  \addplot+[red,solid,mark=none, thick]                                  table[x=ncands,y=jac_instance] {data/obj_cands/pascal2012_val2012_SCG.txt};
	  	  \addplot+[cyan,solid,mark=none, thick]                                 table[x=ncands,y=jac_instance] {data/obj_cands/pascal2012_val2012_CPMC.txt};
	      \addplot+[only marks,red,solid,mark size=1.75,mark=asterisk, thick]    table[x=ncands,y=jac_instance] {data/obj_cands/pascal2012_val2012_RIGOR.txt};
	      \addplot+[only marks,blue,solid,mark=triangle,mark size=1.9, thick]    table[x=ncands,y=jac_instance] {data/obj_cands/pascal2012_val2012_GOP.txt};
	      \addplot+[only marks,cyan,solid,mark=square,mark size=1.5, thick]      table[x=ncands,y=jac_instance] {data/obj_cands/pascal2012_val2012_GLS.txt};
          \addplot+[only marks,magenta,solid,mark=triangle,mark size=1.9, thick] table[x=ncands,y=jac_instance] {data/obj_cands/pascal2012_val2012_ShSh.txt};
	  	  \addplot+[olive,solid,mark=none, thick]                                table[x=ncands,y=jac_instance] {data/obj_cands/pascal2012_val2012_CI.txt};
		  \addplot+[only marks,black,solid,mark=square, mark size=1.5, thick]    table[x=ncands,y=jac_instance] {data/obj_cands/pascal2012_val2012_SeSe.txt};
		  \addplot+[red,dashed,mark=none, thick]                                 table[x=ncands,y=jac_instance] {data/obj_cands/val2012_sf_mUCM_multi_3sc_u_4r_12k_single_multi.txt};
    	  \addplot+[green,dashed,mark=none, thick]                               table[x=ncands,y=jaco] {data/obj_cands/val2012_ucm.txt};
		  \addplot+[black,dashed,mark=none, thick]                               table[x=ncands,y=jac_instance] {data/obj_cands/pascal2012_val2012_QT.txt};
	\end{axis}
   \end{tikzpicture}
\end{minipage}
\begin{minipage}[b]{0.39\linewidth}
\centering
\scriptsize \hspace{5mm}SBD\\[0.5mm]
\begin{tikzpicture}[/pgfplots/width=1.1\linewidth, /pgfplots/height=1.25\linewidth]
    \begin{axis}[ymin=0.25,ymax=0.9,xmin=20,xmax=15000,enlargelimits=false,
        xlabel=Number of proposals,
        ylabel=Jaccard index at instance level ($J_i$),
        font=\scriptsize, grid=both,
        grid style=dotted,
        axis equal image=false,
        ytick={0,0.1,...,1},
        minor ytick={0,0.025,...,1},
        major grid style={white!20!black},
        minor grid style={white!70!black},
        xlabel shift={-2pt},
        ylabel shift={-3pt},
        xmode=log]
		\addplot+[black,solid,mark=none, ultra thick]                          table[x=ncands,y=jac_instance] {data/obj_cands/SBD_val_MCG.txt};
        \addplot+[red,solid,mark=none, thick]                                  table[x=ncands,y=jac_instance] {data/obj_cands/SBD_val_SCG.txt};
	    \addplot+[only marks,blue,solid,mark=triangle,mark size=1.9, thick]    table[x=ncands,y=jac_instance] {data/obj_cands/SBD_val_GOP.txt};
	    \addplot+[only marks,red,solid,mark=asterisk, mark size=1.75, thick]   table[x=ncands,y=jac_instance] {data/obj_cands/SBD_val_RIGOR.txt};
	    \addplot+[only marks,cyan,solid,mark=square,mark size=1.5, thick]      table[x=ncands,y=jac_instance] {data/obj_cands/SBD_val_GLS.txt};
		\addplot+[only marks,black,solid,mark=square, mark size=1.5, thick]    table[x=ncands,y=jac_instance] {data/obj_cands/SBD_val_SeSe.txt};
		\addplot+[black,dashed,mark=none, thick]                               table[x=ncands,y=jac_instance] {data/obj_cands/SBD_val_QT.txt};

	\end{axis}
   \end{tikzpicture}\end{minipage}
\begin{minipage}[b]{0.39\linewidth}
\centering
\scriptsize \hspace{5mm}COCO\\[0.5mm]
\begin{tikzpicture}[/pgfplots/width=1.1\linewidth, /pgfplots/height=1.25\linewidth]
    \begin{axis}[ymin=0.25,ymax=0.9,xmin=20,xmax=15000,enlargelimits=false,
        xlabel=Number of proposals,
        ylabel=Jaccard index at instance level ($J_i$),
        font=\scriptsize, grid=both,
        grid style=dotted,
        axis equal image=false,
        legend pos= outer north east,
        ytick={0,0.1,...,1},
        minor ytick={0,0.025,...,1},
        major grid style={white!20!black},
        minor grid style={white!70!black},
        xlabel shift={-2pt},
        ylabel shift={-3pt},
        xmode=log]
        \addplot[black,solid,mark=none, ultra thick] coordinates {( 1, 0.7)( 1, 0.8)};
        \addplot[red,solid,mark=none, thick] coordinates {( 1, 0.7)( 1, 0.8)};
	    \addplot[cyan,solid,mark=none, thick] coordinates {( 1, 0.7)( 1, 0.8)};
	    \addplot[olive,solid,mark=none, thick] coordinates {( 1, 0.7)( 1, 0.8)};
        \addplot[only marks,blue,solid,mark=triangle,mark size=1.9, thick] coordinates {( 1, 0.7)( 1, 0.8)};
	  	\addplot[only marks,cyan,solid,mark=square,mark size=1.5, thick] coordinates {( 1, 0.7)( 1, 0.8)};
	    \addplot[only marks,red,solid,mark=asterisk, mark size=1.75, thick] coordinates {( 1, 0.7)( 1, 0.8)};
  	    \addplot[only marks,magenta,solid,mark=triangle,mark size=1.9, thick] coordinates {( 1, 0.7)( 1, 0.8)};
	    \addplot[only marks,black,solid,mark=square, mark size=1.5, thick] coordinates {( 1, 0.7)( 1, 0.8)};
	    \addplot[red,dashed,mark=none, thick]  coordinates {( 1, 0.7)( 1, 0.8)};
    	\addplot[green,dashed,mark=none, thick] coordinates {( 1, 0.7)( 1, 0.8)};
		\addplot[black,dashed,mark=none, thick] coordinates {( 1, 0.7)( 1, 0.8)};
		  
		\addlegendentry{MCG-Our}
        \addlegendentry{SCG-Our}
   	    \addlegendentry{CPMC~\cite{Carreira2012b}}
	   \addlegendentry{CI~\cite{Endres2014}}
        \addlegendentry{GOP~\cite{Kraehenbuehl2014}}
		\addlegendentry{GLS~\cite{Rantalankila2014}}
	   \addlegendentry{RIGOR~\cite{Humayun2014}}
	   \addlegendentry{ShSh~\cite{Kim2012}}
	   \addlegendentry{SeSe~\cite{Uijlings2013}}
	   \addlegendentry{MCG-UCM-Our}
	   \addlegendentry{gPb-UCM}
	   \addlegendentry{Quadtree}
	   
	    \addplot+[black,solid,mark=none, ultra thick]                        table[x=ncands,y=jac_instance] {data/obj_cands/COCO_val2014_MCG.txt};
        \addplot+[red,solid,mark=none, thick]                                table[x=ncands,y=jac_instance] {data/obj_cands/COCO_val2014_SCG.txt};
        \addplot+[only marks,blue,solid,mark=triangle,mark size=1.9, thick]  table[x=ncands,y=jac_instance] {data/obj_cands/COCO_val2014_GOP.txt};
	  	\addplot+[only marks,cyan,solid,mark=square,mark size=1.5, thick]    table[x=ncands,y=jac_instance] {data/obj_cands/COCO_val2014_GLS.txt};
		\addplot+[only marks,black,solid,mark=square, mark size=1.5, thick]  table[x=ncands,y=jac_instance] {data/obj_cands/COCO_val2014_SeSe.txt};
	    \addplot+[only marks,red,solid,mark=asterisk, mark size=1.75, thick] table[x=ncands,y=jac_instance] {data/obj_cands/COCO_val2014_RIGOR.txt};
		\addplot+[black,dashed,mark=none, thick]                             table[x=ncands,y=jac_instance] {data/obj_cands/COCO_val2014_QT.txt};

	\end{axis}
   \end{tikzpicture}\end{minipage}}
   \caption{\textbf{Object Proposals: Jaccard index at instance level.} Results on SegVOC12, SBD, and COCO.}
   \label{fig:J_i}
\end{figure*}

The improvement of considering the combination of all 1-region proposals (\ref{fig:train:singletons}) from the 3 scales
and the MCG-UCM-Our with respect to the raw MCG-UCM-Our (\ref{ours-multi-singletons}) is significant,
which corroborates the gain in diversity obtained from hierarchies at different scales.
In turn, the addition of 2- and 3-region proposals (\ref{fig:train:pairs}~and~\ref{fig:train:triplets})
noticeably improves the achievable quality. 
This shows that hierarchies do not get full objects in single regions, which makes sense given that
they are built using low-level features only.
The improvement when adding 4-tuples (\ref{fig:train:4tuples}) is marginal at the number of proposals we are considering.
When analyzing the equal distribution of proposals from the four scales (\ref{fig:train:equal}), we see that the less proposals we consider, the more relevant the Pareto optimization becomes.
At the selected working point, the gain of the Pareto optimization is 2 points.

\begin{figure}
\centering
\resizebox{0.8\linewidth}{!}{\begin{tikzpicture}[/pgfplots/width=\linewidth, /pgfplots/height=0.8\linewidth, /pgfplots/legend pos=south east]
    \begin{axis}[ymin=0,ymax=100,xmin=500,xmax=100000,enlargelimits=false,area style,
        xlabel=Number of proposals,
        ylabel=Region distribution percentage,
		font=\footnotesize,
        xlabel shift={-2pt},
        ylabel shift={-5pt},
        xmode=log,
        legend style={at={(0.03,0.3)},anchor=west}]
        
            \addplot+[blue!80!white,fill=blue!80!white,mark=none,solid]  table[x=n_regs,y=scale_2.00] {data/pareto/pareto_reg_distr.txt};
    \addlegendentry[font=\scriptsize]{Scale 2.0}
    \addplot+[blue!60!white,fill=blue!60!white,mark=none,solid]  table[x=n_regs,y=scale_1.00] {data/pareto/pareto_reg_distr.txt};
    \addlegendentry[font=\scriptsize]{Scale 1.0}
    \addplot+[blue!50!white,fill=blue!50!white,mark=none,solid]  table[x=n_regs,y=scale_0.50] {data/pareto/pareto_reg_distr.txt};
    \addlegendentry[font=\scriptsize]{Scale 0.5}
    \addplot+[blue!40!white,fill=blue!40!white,mark=none,solid]  table[x=n_regs,y=multi] {data/pareto/pareto_reg_distr.txt};
    \addlegendentry[font=\scriptsize]{Multi-Scale}

    \addplot+[forget plot,fill,blue!80!white]  table[x=n_regs,y=scale_2.00] {data/pareto/pareto_reg_distr.txt} \closedcycle;
    \addplot+[forget plot,fill,blue!60!white]  table[x=n_regs,y=scale_1.00] {data/pareto/pareto_reg_distr.txt} \closedcycle;
    \addplot+[forget plot,fill,blue!50!white]  table[x=n_regs,y=scale_0.50] {data/pareto/pareto_reg_distr.txt} \closedcycle;
     \addplot+[forget plot,fill,blue!40!white]  table[x=n_regs,y=multi] {data/pareto/pareto_reg_distr.txt} \closedcycle;
    \end{axis}
\end{tikzpicture}}
   \caption{Region distribution learnt by the Pareto front optimization on SegVOC12.}
   \label{fig:reg_dist}
\end{figure}
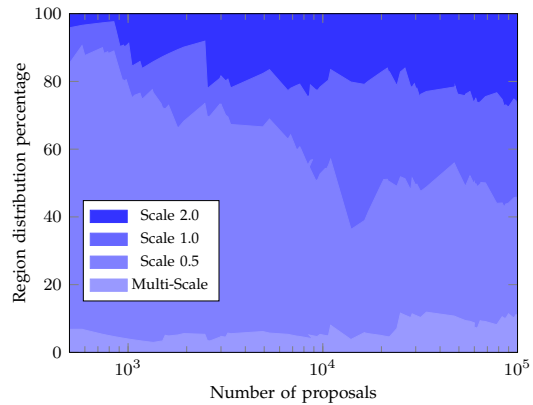

Figure~\ref{fig:reg_dist} shows the distribution of proposals from each of the scales combined in the Pareto front.
We see that the coarse scale (0.5) is the most \textit{picked} at low number of proposals,
and the rest come into play when increasing their number,
since one can afford more detailed proposals.
The multi-scale hierarchy is the one with less weight, since it is created from the other three.

\paragraph*{\textbf{Pareto selection and ranking}}
Back to Figure~\ref{fig:obj_cands_train}, the red asterisk
(\ref{marker:jacc_train:sel_pareto}) marks the selected configuration
$\left\{\overline{N}_1,\dots,\overline{N}_R\right\}$ in the Pareto front (black triangle in Figure~\ref{fig:pareto}),
which is selected at a practical level of proposals.
The red plus sign (\ref{marker:jacc_train:filt_pareto}) represents the set of proposals
after removing those duplicate proposals whose overlap leads to a Jaccard higher than 0.95.
The proposals at this point are the ones that are ranked by the learnt regressor (\ref{marker:jacc_train:regressed}).

At test time (right-hand plot), we directly combine the learnt $\left\{\overline{N}_1,\dots,\overline{N}_R\right\}$ proposals from each ranked list. Note that the Pareto optimization does not overfit, given the similar result in the training and validation datasets.
We then remove duplicates and rank the results.
In this case, note the difference between the regressed result in the training and validation sets,
which reflects overfitting, but despite this we found it beneficial with respect to the non-regressed
result.

In the validation set of SegVOC12, the full set of proposals (i.e., combining the full 16 lists)
would contain millions of proposals per image.
The multiscale combinatorial grouping allows us to reduce the number of proposals to $5\,086$ 
with a very high achievable $J_i$ of $0.81$ (\ref{marker:jacc_train:filt_pareto}).
The regressed ranking (\ref{marker:jacc_train:regressed}) allows us to further reduce the number of proposals below this point.


\paragraph*{\textbf{Segmented Proposals: Comparison with State of the Art}}
We first compare our results against those methods that produce segmented object proposals~\cite{Kraehenbuehl2014,Rantalankila2014,Humayun2014,Uijlings2013,Alexe2012,Kim2012,Carreira2012b,Endres2014},
using the implementations from the respective authors.
We train MCG on the training set of SegVOC12, and we use the learnt parameters on the validation sets of SegVOC12, SBD, and COCO.

Figure~\ref{fig:J_i} shows the achievable quality at instance level ($J_i$)
of all methods on the validation set of SegVOC12, SBD, and COCO.
We plot the raw regions of MCG-UCM-Our, gPb-UCM, and QuadTree as baselines where available.
We also evaluate a faster single-scale version of MCG (\textit{Single-scale Combinatorial Grouping - SCG}),
which takes the hierarchy at the native scale only and combines up to 4 regions per proposal.   
This approach decreases the computational load one order of magnitude while keeping competitive results.

\begin{figure*}
\scalebox{0.76}{\hspace{-2mm}\begin{minipage}[b]{0.39\linewidth}
\centering
\scriptsize \hspace{5mm}PASCAL SegVOC12\\[1mm]
\begin{tikzpicture}[/pgfplots/width=1.1\linewidth, /pgfplots/height=1.27\linewidth]
    \begin{axis}[ymin=0,ymax=1,xmin=20,xmax=15000,enlargelimits=false,
        xlabel=Number of proposals,
        ylabel=Recall,
        font=\scriptsize, grid=both,
        grid style=dotted,
        axis equal image=false,
        ytick={0,0.1,...,1},
        minor ytick={0,0.025,...,1},
        major grid style={white!20!black},
        minor grid style={white!70!black},
        xlabel shift={-2pt},
        ylabel shift={-3pt},
        xmode=log,
        legend style={
		at={(2.39,0.98)},
		fill=white,
		fill opacity=1,
		anchor=north west}]
        \addplot+[black,solid,mark=none, ultra thick]                          table[x=ncands,y=rec_at_0.5] {data/obj_cands/pascal2012_val2012_MCG.txt};
        \addplot+[red,solid,mark=none, thick]                                  table[x=ncands,y=rec_at_0.5] {data/obj_cands/pascal2012_val2012_SCG.txt};
        \addplot+[only marks,blue,solid,mark=triangle,mark size=1.9, thick]    table[x=ncands,y=rec_at_0.5] {data/obj_cands/pascal2012_val2012_GOP.txt};
	  	\addplot+[only marks,cyan,solid,mark=square,mark size=1.5, thick]      table[x=ncands,y=rec_at_0.5] {data/obj_cands/pascal2012_val2012_GLS.txt};
	    \addplot+[cyan,solid,mark=none, thick]                                 table[x=ncands,y=rec_at_0.5] {data/obj_cands/pascal2012_val2012_CPMC.txt};
	  	\addplot+[olive,solid,mark=none, thick]                                table[x=ncands,y=rec_at_0.5] {data/obj_cands/pascal2012_val2012_CI.txt};
		\addplot+[only marks,red,solid,mark=asterisk, mark size=1.75, thick]   table[x=ncands,y=rec_at_0.5] {data/obj_cands/pascal2012_val2012_RIGOR.txt};
  	    \addplot+[only marks,magenta,solid,mark=triangle,mark size=1.9, thick] table[x=ncands,y=rec_at_0.5] {data/obj_cands/pascal2012_val2012_ShSh.txt};
		\addplot+[only marks,black,solid,mark=square, mark size=1.5, thick]    table[x=ncands,y=rec_at_0.5] {data/obj_cands/pascal2012_val2012_SeSe.txt};
	    \addplot+[black,dashed,mark=none,thick]                                table[x=ncands,y=rec_at_0.5] {data/obj_cands/pascal2012_val2012_QT.txt};

		\addplot+[black,solid,mark=none, ultra thick]                          table[x=ncands,y=rec_at_0.7] {data/obj_cands/pascal2012_val2012_MCG.txt};
        \addplot+[red,solid,mark=none, thick]                                  table[x=ncands,y=rec_at_0.7] {data/obj_cands/pascal2012_val2012_SCG.txt};
        \addplot+[only marks,blue,solid,mark=triangle,mark size=1.9, thick]    table[x=ncands,y=rec_at_0.7] {data/obj_cands/pascal2012_val2012_GOP.txt};
	  	\addplot+[only marks,cyan,solid,mark=square,mark size=1.5, thick]      table[x=ncands,y=rec_at_0.7] {data/obj_cands/pascal2012_val2012_GLS.txt};
        \addplot+[cyan,solid,mark=none, thick]                                 table[x=ncands,y=rec_at_0.7] {data/obj_cands/pascal2012_val2012_CPMC.txt};
	  	\addplot+[olive,solid,mark=none, thick]                                table[x=ncands,y=rec_at_0.7] {data/obj_cands/pascal2012_val2012_CI.txt};
		\addplot+[only marks,red,solid,mark=asterisk, mark size=1.75, thick]   table[x=ncands,y=rec_at_0.7] {data/obj_cands/pascal2012_val2012_RIGOR.txt};
  	    \addplot+[only marks,magenta,solid,mark=triangle,mark size=1.9, thick] table[x=ncands,y=rec_at_0.7] {data/obj_cands/pascal2012_val2012_ShSh.txt};
		\addplot+[only marks,black,solid,mark=square, mark size=1.5, thick]    table[x=ncands,y=rec_at_0.7] {data/obj_cands/pascal2012_val2012_SeSe.txt};
		\addplot+[black,dashed,mark=none,thick]                                table[x=ncands,y=rec_at_0.7] {data/obj_cands/pascal2012_val2012_QT.txt};
        
        \addplot+[black,solid,mark=none, ultra thick]                          table[x=ncands,y=rec_at_0.85] {data/obj_cands/pascal2012_val2012_MCG.txt};
        \addplot+[red,solid,mark=none, thick]                                  table[x=ncands,y=rec_at_0.85] {data/obj_cands/pascal2012_val2012_SCG.txt};
        \addplot+[only marks,blue,solid,mark=triangle,mark size=1.9, thick]    table[x=ncands,y=rec_at_0.85] {data/obj_cands/pascal2012_val2012_GOP.txt};
	  	\addplot+[only marks,cyan,solid,mark=square,mark size=1.5, thick]      table[x=ncands,y=rec_at_0.85] {data/obj_cands/pascal2012_val2012_GLS.txt};
	    \addplot+[cyan,solid,mark=none, thick]                                 table[x=ncands,y=rec_at_0.85] {data/obj_cands/pascal2012_val2012_CPMC.txt};
	  	\addplot+[olive,solid,mark=none, thick]                                table[x=ncands,y=rec_at_0.85] {data/obj_cands/pascal2012_val2012_CI.txt};
		\addplot+[only marks,red,solid,mark=asterisk, mark size=1.75, thick]   table[x=ncands,y=rec_at_0.85] {data/obj_cands/pascal2012_val2012_RIGOR.txt};
  	    \addplot+[only marks,magenta,solid,mark=triangle,mark size=1.9, thick] table[x=ncands,y=rec_at_0.85] {data/obj_cands/pascal2012_val2012_ShSh.txt};
		\addplot+[only marks,black,solid,mark=square, mark size=1.5, thick]    table[x=ncands,y=rec_at_0.85] {data/obj_cands/pascal2012_val2012_SeSe.txt};
		\addplot+[black,dashed,mark=none,thick]                                table[x=ncands,y=rec_at_0.85] {data/obj_cands/pascal2012_val2012_QT.txt};

	   \node[draw,fill=white,solid,anchor=north west] at (axis cs:20,0.53) {$J\!=\!0.5$};
	   \node[draw,fill=white,solid,anchor=north west] at (axis cs:20,0.33) {$J\!=\!0.7$};
	   \node[draw,fill=white,solid,anchor=north west] at (axis cs:20,0.15) {$J\!=\!0.85$};
	\end{axis}
   \end{tikzpicture}
\end{minipage}
\begin{minipage}[b]{0.39\linewidth}
\centering
\scriptsize \hspace{5mm}SBD\\[1mm]
\begin{tikzpicture}[/pgfplots/width=1.1\linewidth, /pgfplots/height=1.27\linewidth]
    \begin{axis}[ymin=0,ymax=1,xmin=20,xmax=15000,enlargelimits=false,
        xlabel=Number of proposals,
        ylabel=Recall,
        font=\scriptsize, grid=both,
        grid style=dotted,
        axis equal image=false,
        ytick={0,0.1,...,1},
        minor ytick={0,0.025,...,1},
        major grid style={white!20!black},
        minor grid style={white!70!black},
        xlabel shift={-2pt},
        ylabel shift={-3pt},
        xmode=log]
        \addplot+[black,solid,mark=none, ultra thick]                          table[x=ncands,y=rec_at_0.5] {data/obj_cands/SBD_val_MCG.txt};
        \addplot+[red,solid,mark=none, thick]                                  table[x=ncands,y=rec_at_0.5] {data/obj_cands/SBD_val_SCG.txt};
        \addplot+[only marks,blue,solid,mark=triangle,mark size=1.9, thick]    table[x=ncands,y=rec_at_0.5] {data/obj_cands/SBD_val_GOP.txt};
	  	\addplot+[only marks,cyan,solid,mark=square,mark size=1.5, thick]      table[x=ncands,y=rec_at_0.5] {data/obj_cands/SBD_val_GLS.txt};
		\addplot+[only marks,red,solid,mark=asterisk, mark size=1.75, thick]   table[x=ncands,y=rec_at_0.5] {data/obj_cands/SBD_val_RIGOR.txt};
		\addplot+[only marks,black,solid,mark=square, mark size=1.5, thick]    table[x=ncands,y=rec_at_0.5] {data/obj_cands/SBD_val_SeSe.txt};
		\addplot+[black,dashed,mark=none,thick]                                table[x=ncands,y=rec_at_0.5] {data/obj_cands/SBD_val_QT.txt};

		\addplot+[black,solid,mark=none, ultra thick]                          table[x=ncands,y=rec_at_0.7] {data/obj_cands/SBD_val_MCG.txt};
        \addplot+[red,solid,mark=none, thick]                                  table[x=ncands,y=rec_at_0.7] {data/obj_cands/SBD_val_SCG.txt};
        \addplot+[only marks,blue,solid,mark=triangle,mark size=1.9, thick]    table[x=ncands,y=rec_at_0.7] {data/obj_cands/SBD_val_GOP.txt};
	  	\addplot+[only marks,cyan,solid,mark=square,mark size=1.5, thick]      table[x=ncands,y=rec_at_0.7] {data/obj_cands/SBD_val_GLS.txt};
		\addplot+[only marks,red,solid,mark=asterisk, mark size=1.75, thick]   table[x=ncands,y=rec_at_0.7] {data/obj_cands/SBD_val_RIGOR.txt};
		\addplot+[only marks,black,solid,mark=square, mark size=1.5, thick]    table[x=ncands,y=rec_at_0.7] {data/obj_cands/SBD_val_SeSe.txt};
		\addplot+[black,dashed,mark=none,thick]                                table[x=ncands,y=rec_at_0.7] {data/obj_cands/SBD_val_QT.txt};

        \addplot+[black,solid,mark=none, ultra thick]                          table[x=ncands,y=rec_at_0.85] {data/obj_cands/SBD_val_MCG.txt};
        \addplot+[red,solid,mark=none, thick]                                  table[x=ncands,y=rec_at_0.85] {data/obj_cands/SBD_val_SCG.txt};
        \addplot+[only marks,blue,solid,mark=triangle,mark size=1.9, thick]    table[x=ncands,y=rec_at_0.85] {data/obj_cands/SBD_val_GOP.txt};
	  	\addplot+[only marks,cyan,solid,mark=square,mark size=1.5, thick]      table[x=ncands,y=rec_at_0.85] {data/obj_cands/SBD_val_GLS.txt};
		\addplot+[only marks,red,solid,mark=asterisk, mark size=1.75, thick]   table[x=ncands,y=rec_at_0.85] {data/obj_cands/SBD_val_RIGOR.txt};
		\addplot+[only marks,black,solid,mark=square, mark size=1.5, thick]    table[x=ncands,y=rec_at_0.85] {data/obj_cands/SBD_val_SeSe.txt};
		\addplot+[black,dashed,mark=none,thick]                                table[x=ncands,y=rec_at_0.85] {data/obj_cands/SBD_val_QT.txt};

	   \node[draw,fill=white,solid,anchor=north west] at (axis cs:20,0.5) {$J=0.5$};
	   \node[draw,fill=white,solid,anchor=north west] at (axis cs:20,0.3) {$J=0.7$};
	   \node[draw,fill=white,solid,anchor=north west] at (axis cs:20,0.12) {$J=0.85$};
	\end{axis}
   \end{tikzpicture}\end{minipage}
\begin{minipage}[b]{0.39\linewidth}
\centering
\scriptsize \hspace{5mm}COCO\\[1mm]
\begin{tikzpicture}[/pgfplots/width=1.1\linewidth, /pgfplots/height=1.27\linewidth]
    \begin{axis}[ymin=0,ymax=1,xmin=20,xmax=15000,enlargelimits=false,
        xlabel=Number of proposals,
        ylabel=Recall,
        font=\scriptsize, grid=both,
        grid style=dotted,
        axis equal image=false,
        legend pos= outer north east,
        ytick={0,0.1,...,1},
        minor ytick={0,0.025,...,1},
        major grid style={white!20!black},
        minor grid style={white!70!black},
        xlabel shift={-2pt},
        ylabel shift={-3pt},
        xmode=log]
        \addplot[black,solid,mark=none, ultra thick]                          coordinates {( 1, 0.7)( 1, 0.8)};
        \label{fig:recall:mcg}
        \addplot[red,solid,mark=none, thick]                                  coordinates {( 1, 0.7)( 1, 0.8)};
        \label{fig:recall:scg}
	    \addplot[cyan,solid,mark=none, thick]                                 coordinates {( 1, 0.7)( 1, 0.8)};
	    \label{fig:recall:cpmc}
	    \addplot[olive,solid,mark=none, thick]                                coordinates {( 1, 0.7)( 1, 0.8)};
	    \label{fig:recall:ci}
        \addplot[only marks,blue,solid,mark=triangle,mark size=1.9, thick]    coordinates {( 1, 0.7)( 1, 0.8)};
        \label{fig:recall:gop}
	  	\addplot[only marks,cyan,solid,mark=square,mark size=1.5, thick]      coordinates {( 1, 0.7)( 1, 0.8)};
		\label{fig:recall:gls}
	    \addplot[only marks,red,solid,mark=asterisk, mark size=1.75, thick]   coordinates {( 1, 0.7)( 1, 0.8)};
	    \label{fig:recall:rigor}
  	    \addplot[only marks,magenta,solid,mark=triangle,mark size=1.9, thick] coordinates {( 1, 0.7)( 1, 0.8)};
	    \label{fig:recall:shsh}
	    \addplot[only marks,black,solid,mark=square, mark size=1.5, thick]    coordinates {( 1, 0.7)( 1, 0.8)};
	    \label{fig:recall:sese}
	    \addplot[black,dashed,mark=none,thick]                                coordinates {( 1, 0.7)( 1, 0.8)};
        \label{fig:recall:qt}
        
		\addlegendentry{MCG-Our}
        \addlegendentry{SCG-Our}
        \addlegendentry{CPMC~\cite{Carreira2012b}}
	    \addlegendentry{CI~\cite{Endres2014}}
        \addlegendentry{GOP~\cite{Kraehenbuehl2014}}
		\addlegendentry{GLS~\cite{Rantalankila2014}}
	    \addlegendentry{RIGOR~\cite{Humayun2014}}
	    \addlegendentry{ShSh~\cite{Kim2012}}
	    \addlegendentry{SeSe~\cite{Uijlings2013}}
	   	\addlegendentry{Quadtree}

        \addplot+[black,solid,mark=none, ultra thick]                          table[x=ncands,y=rec_at_0.5] {data/obj_cands/COCO_val2014_MCG.txt};
        \addplot+[red,solid,mark=none, thick]                                  table[x=ncands,y=rec_at_0.5] {data/obj_cands/COCO_val2014_SCG.txt};
        \addplot+[only marks,blue,solid,mark=triangle,mark size=1.9, thick]    table[x=ncands,y=rec_at_0.5] {data/obj_cands/COCO_val2014_GOP.txt};
	  	\addplot+[only marks,cyan,solid,mark=square,mark size=1.5, thick]      table[x=ncands,y=rec_at_0.5] {data/obj_cands/COCO_val2014_GLS.txt};
	    \addplot+[only marks,red,solid,mark=asterisk, mark size=1.75, thick]   table[x=ncands,y=rec_at_0.5] {data/obj_cands/COCO_val2014_RIGOR.txt};
	    \addplot+[only marks,black,solid,mark=square, mark size=1.5, thick]    table[x=ncands,y=rec_at_0.5] {data/obj_cands/COCO_val2014_SeSe.txt};
		\addplot+[black,dashed,mark=none,thick]                                table[x=ncands,y=rec_at_0.5] {data/obj_cands/COCO_val2014_QT.txt};

		\addplot+[black,solid,mark=none, ultra thick]                          table[x=ncands,y=rec_at_0.7] {data/obj_cands/COCO_val2014_MCG.txt};
        \addplot+[red,solid,mark=none, thick]                                  table[x=ncands,y=rec_at_0.7] {data/obj_cands/COCO_val2014_SCG.txt};
        \addplot+[only marks,blue,solid,mark=triangle,mark size=1.9, thick]    table[x=ncands,y=rec_at_0.7] {data/obj_cands/COCO_val2014_GOP.txt};
	  	\addplot+[only marks,cyan,solid,mark=square,mark size=1.5, thick]      table[x=ncands,y=rec_at_0.7] {data/obj_cands/COCO_val2014_GLS.txt};
	    \addplot+[only marks,red,solid,mark=asterisk, mark size=1.75, thick]   table[x=ncands,y=rec_at_0.7] {data/obj_cands/COCO_val2014_RIGOR.txt};
	    \addplot+[only marks,black,solid,mark=square, mark size=1.5, thick]    table[x=ncands,y=rec_at_0.7] {data/obj_cands/COCO_val2014_SeSe.txt};
		\addplot+[black,dashed,mark=none,thick]                                table[x=ncands,y=rec_at_0.7] {data/obj_cands/COCO_val2014_QT.txt};

        \addplot+[black,solid,mark=none, ultra thick]                          table[x=ncands,y=rec_at_0.85] {data/obj_cands/COCO_val2014_MCG.txt};
        \addplot+[red,solid,mark=none, thick]                                  table[x=ncands,y=rec_at_0.85] {data/obj_cands/COCO_val2014_SCG.txt};
        \addplot+[only marks,blue,solid,mark=triangle,mark size=1.9, thick]    table[x=ncands,y=rec_at_0.85] {data/obj_cands/COCO_val2014_GOP.txt};
	  	\addplot+[only marks,cyan,solid,mark=square,mark size=1.5, thick]      table[x=ncands,y=rec_at_0.85] {data/obj_cands/COCO_val2014_GLS.txt};
	    \addplot+[only marks,red,solid,mark=asterisk, mark size=1.75, thick]   table[x=ncands,y=rec_at_0.85] {data/obj_cands/COCO_val2014_RIGOR.txt};
	    \addplot+[only marks,black,solid,mark=square, mark size=1.5, thick]    table[x=ncands,y=rec_at_0.85] {data/obj_cands/COCO_val2014_SeSe.txt};
		\addplot+[black,dashed,mark=none,thick]                                table[x=ncands,y=rec_at_0.85] {data/obj_cands/COCO_val2014_QT.txt};

	   \node[draw,fill=white,solid,anchor=north west] at (axis cs:20,0.23) {$J=0.5$};
	   \node[draw,fill=white,solid,anchor=north west] at (axis cs:20,0.12) {$J=0.7$};
	   \node[draw,fill=white,solid,anchor=north west] at (axis cs:20,0.05) {$J=0.85$};
	\end{axis}
   \end{tikzpicture}\end{minipage}}
   \caption{\textbf{Segmented Object Proposals: Recall at different Jaccard levels.} Percentage of annotated objects for which there is a proposal whose overlap with the segmented ground-truth shapes (not boxes) is above $J=0.5$, $J=0.7$, and $J=0.85$, for different number of proposals per image. Results on SegVOC12, SBD, and COCO.}
   \label{fig:recall_segmented}
\end{figure*}
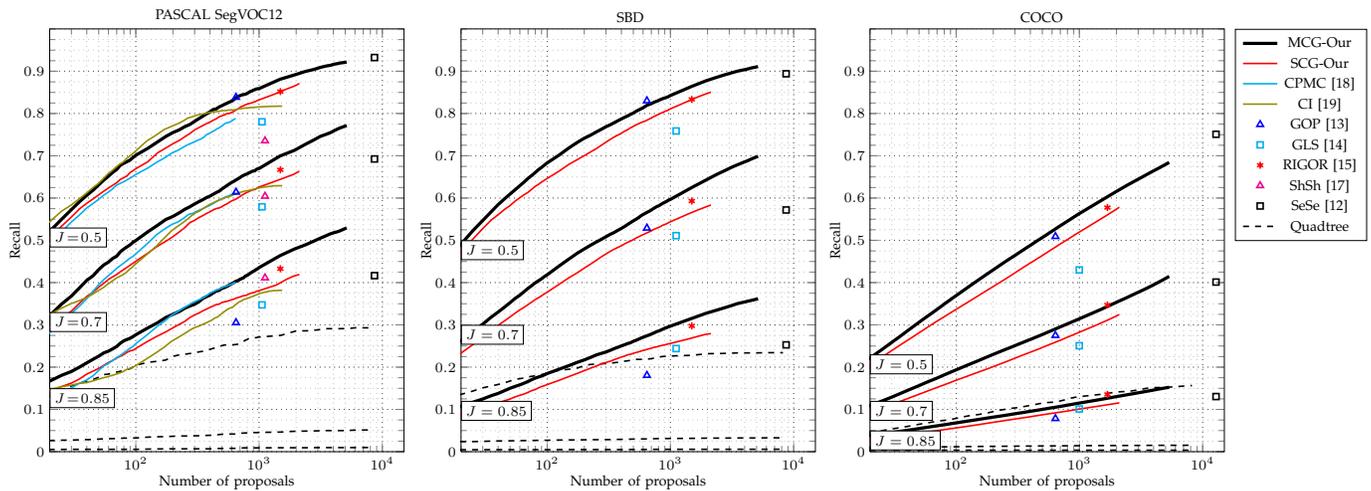

MCG proposals (\ref{fig:recall:mcg}) significantly outperform the state-of-the-art at all regimes.
The bigger the database is, the better MCG results are with respect to the rest, 
which shows that our techniques better generalize to unseen images
(recall that MCG is trained only in SegVOC12).

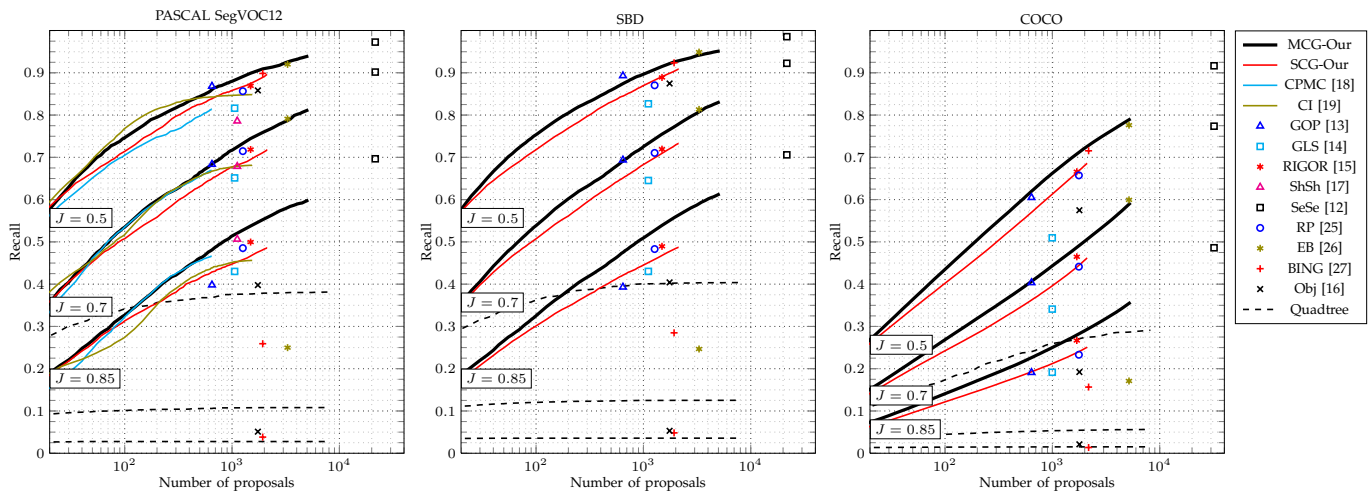
\begin{figure*}
\scalebox{0.76}{\hspace{-2mm}\begin{minipage}[b]{0.39\linewidth}
\centering
\scriptsize \hspace{5mm}PASCAL SegVOC12\\[1mm]
\begin{tikzpicture}[/pgfplots/width=1.1\linewidth, /pgfplots/height=1.27\linewidth]
    \begin{axis}[ymin=0,ymax=1,xmin=20,xmax=40000,enlargelimits=false,
        xlabel=Number of proposals,
        ylabel=Recall,
        font=\scriptsize, grid=both,
        grid style=dotted,
        axis equal image=false,
        ytick={0,0.1,...,1},
        minor ytick={0,0.025,...,1},
        major grid style={white!20!black},
        minor grid style={white!70!black},
        xlabel shift={-2pt},
        ylabel shift={-3pt},
        xmode=log,
        legend style={
		at={(2.39,0.98)},
		fill=white,
		fill opacity=1,
		anchor=north west}]
        \addplot+[black,solid,mark=none, ultra thick]                            table[x=ncands,y=rec_at_0.5] {data/obj_cands/pascal2012_val2012_MCG_boxes.txt};

        \addplot+[red,solid,mark=none, thick]                                    table[x=ncands,y=rec_at_0.5] {data/obj_cands/pascal2012_val2012_SCG_boxes.txt};
        \addplot+[only marks,blue,solid,mark=triangle,mark size=1.9, thick]      table[x=ncands,y=rec_at_0.5] {data/obj_cands/pascal2012_val2012_GOP_boxes.txt};
	  	\addplot+[only marks,cyan,solid,mark=square,mark size=1.5, thick]        table[x=ncands,y=rec_at_0.5] {data/obj_cands/pascal2012_val2012_GLS_boxes.txt};
	    \addplot+[cyan,solid,mark=none, thick]                                   table[x=ncands,y=rec_at_0.5] {data/obj_cands/pascal2012_val2012_CPMC_boxes.txt};
	    \addplot+[olive,solid,mark=none, thick]                                  table[x=ncands,y=rec_at_0.5] {data/obj_cands/pascal2012_val2012_CI_boxes.txt};
		\addplot+[only marks,red,solid,mark=asterisk, mark size=1.75, thick]     table[x=ncands,y=rec_at_0.5] {data/obj_cands/pascal2012_val2012_RIGOR_boxes.txt};
  	    \addplot+[only marks,magenta,solid,mark=triangle,mark size=1.9, thick]  table[x=ncands,y=rec_at_0.5] {data/obj_cands/pascal2012_val2012_ShSh_boxes.txt};
   	    \addplot+[blue,only marks,solid,mark=o, mark size=1.6, thick]            table[x=ncands,y=rec_at_0.5] {data/obj_cands/pascal2012_val2012_RP_boxes.txt};
		\addplot+[only marks,black,solid,mark=square, mark size=1.5, thick]      table[x=ncands,y=rec_at_0.5] {data/obj_cands/pascal2012_val2012_SeSe_boxes.txt};
		\addplot+[only marks,olive  ,solid,mark=asterisk, mark size=1.75, thick] table[x=ncands,y=rec_at_0.5] {data/obj_cands/pascal2012_val2012_EB_boxes.txt};
	    \addplot+[only marks,red    ,solid,mark=+, mark size=1.75, thick]        table[x=ncands,y=rec_at_0.5] {data/obj_cands/pascal2012_val2012_BING_boxes.txt};
	    \addplot+[only marks,black  ,solid,mark=x, thick]                        table[x=ncands,y=rec_at_0.5] {data/obj_cands/pascal2012_val2012_Obj_boxes.txt};
	    \addplot+[black,dashed,mark=none,thick]                                  table[x=ncands,y=rec_at_0.5] {data/obj_cands/pascal2012_val2012_QT_boxes.txt};

		\addplot+[black,solid,mark=none, ultra thick]                            table[x=ncands,y=rec_at_0.7] {data/obj_cands/pascal2012_val2012_MCG_boxes.txt};
        \addplot+[red,solid,mark=none, thick]                                    table[x=ncands,y=rec_at_0.7] {data/obj_cands/pascal2012_val2012_SCG_boxes.txt};
        \addplot+[only marks,blue,solid,mark=triangle,mark size=1.9, thick]      table[x=ncands,y=rec_at_0.7] {data/obj_cands/pascal2012_val2012_GOP_boxes.txt};
	  	\addplot+[only marks,cyan,solid,mark=square,mark size=1.5, thick]        table[x=ncands,y=rec_at_0.7] {data/obj_cands/pascal2012_val2012_GLS_boxes.txt};
	    \addplot+[cyan,solid,mark=none, thick]                                   table[x=ncands,y=rec_at_0.7] {data/obj_cands/pascal2012_val2012_CPMC_boxes.txt};
	  	\addplot+[olive,solid,mark=none, thick]                                  table[x=ncands,y=rec_at_0.7] {data/obj_cands/pascal2012_val2012_CI_boxes.txt};
		\addplot+[only marks,red,solid,mark=asterisk, mark size=1.75, thick]     table[x=ncands,y=rec_at_0.7] {data/obj_cands/pascal2012_val2012_RIGOR_boxes.txt};
  	    \addplot+[only marks,magenta,solid,mark=triangle,mark size=1.9, thick]  table[x=ncands,y=rec_at_0.7] {data/obj_cands/pascal2012_val2012_ShSh_boxes.txt};
   	    \addplot+[blue,only marks,solid,mark=o, mark size=1.6, thick]            table[x=ncands,y=rec_at_0.7] {data/obj_cands/pascal2012_val2012_RP_boxes.txt};
		\addplot+[only marks,black,solid,mark=square, mark size=1.5, thick]      table[x=ncands,y=rec_at_0.7] {data/obj_cands/pascal2012_val2012_SeSe_boxes.txt};
	    \addplot+[only marks,olive  ,solid,mark=asterisk, mark size=1.75, thick] table[x=ncands,y=rec_at_0.7] {data/obj_cands/pascal2012_val2012_EB_boxes.txt};
	    \addplot+[only marks,red    ,solid,mark=+, mark size=1.75, thick]        table[x=ncands,y=rec_at_0.7] {data/obj_cands/pascal2012_val2012_BING_boxes.txt};
	    \addplot+[only marks,black  ,solid,mark=x, thick]                        table[x=ncands,y=rec_at_0.7] {data/obj_cands/pascal2012_val2012_Obj_boxes.txt};
	    \addplot+[black,dashed,mark=none,thick]                                  table[x=ncands,y=rec_at_0.7] {data/obj_cands/pascal2012_val2012_QT_boxes.txt};

        \addplot+[black,solid,mark=none, ultra thick]                            table[x=ncands,y=rec_at_0.85] {data/obj_cands/pascal2012_val2012_MCG_boxes.txt};
        \addplot+[red,solid,mark=none, thick]                                    table[x=ncands,y=rec_at_0.85] {data/obj_cands/pascal2012_val2012_SCG_boxes.txt};
        \addplot+[only marks,blue,solid,mark=triangle,mark size=1.9, thick]      table[x=ncands,y=rec_at_0.85] {data/obj_cands/pascal2012_val2012_GOP_boxes.txt};
	  	\addplot+[only marks,cyan,solid,mark=square,mark size=1.5, thick]        table[x=ncands,y=rec_at_0.85] {data/obj_cands/pascal2012_val2012_GLS_boxes.txt};
	    \addplot+[cyan,solid,mark=none, thick]                                   table[x=ncands,y=rec_at_0.85] {data/obj_cands/pascal2012_val2012_CPMC_boxes.txt};
	  	\addplot+[olive,solid,mark=none, thick]                                  table[x=ncands,y=rec_at_0.85] {data/obj_cands/pascal2012_val2012_CI_boxes.txt};
		\addplot+[only marks,red,solid,mark=asterisk, mark size=1.75, thick]     table[x=ncands,y=rec_at_0.85] {data/obj_cands/pascal2012_val2012_RIGOR_boxes.txt};
  	    \addplot+[only marks,magenta,solid,mark=triangle,mark size=1.9, thick]  table[x=ncands,y=rec_at_0.85] {data/obj_cands/pascal2012_val2012_ShSh_boxes.txt};
   	    \addplot+[blue,only marks   ,solid,mark size=1.6,mark=o, thick]          table[x=ncands,y=rec_at_0.85] {data/obj_cands/pascal2012_val2012_RP_boxes.txt};
		\addplot+[only marks,black  ,solid,mark=square, mark size=1.5, thick]    table[x=ncands,y=rec_at_0.85] {data/obj_cands/pascal2012_val2012_SeSe_boxes.txt};
	    \addplot+[only marks,olive  ,solid,mark=asterisk, mark size=1.75, thick] table[x=ncands,y=rec_at_0.85] {data/obj_cands/pascal2012_val2012_EB_boxes.txt};
	    \addplot+[only marks,red    ,solid,mark=+, mark size=1.75, thick]        table[x=ncands,y=rec_at_0.85] {data/obj_cands/pascal2012_val2012_BING_boxes.txt};
	    \addplot+[only marks,black  ,solid,mark=x, thick]                        table[x=ncands,y=rec_at_0.85] {data/obj_cands/pascal2012_val2012_Obj_boxes.txt};
	    \addplot+[black,dashed,mark=none,thick]                                  table[x=ncands,y=rec_at_0.85] {data/obj_cands/pascal2012_val2012_QT_boxes.txt};

	   \node[draw,fill=white,solid,anchor=north west] at (axis cs:20,0.58) {$J=0.5$};
	   \node[draw,fill=white,solid,anchor=north west] at (axis cs:20,0.37) {$J=0.7$};
	   \node[draw,fill=white,solid,anchor=north west] at (axis cs:20,0.20) {$J=0.85$};
	\end{axis}
   \end{tikzpicture}
\end{minipage}
\begin{minipage}[b]{0.39\linewidth}
\centering
\scriptsize \hspace{5mm}SBD\\[1mm]
\begin{tikzpicture}[/pgfplots/width=1.1\linewidth, /pgfplots/height=1.27\linewidth]
    \begin{axis}[ymin=0,ymax=1,xmin=20,xmax=40000,enlargelimits=false,
        xlabel=Number of proposals,
        ylabel=Recall,
        font=\scriptsize, grid=both,
        grid style=dotted,
        axis equal image=false,
        ytick={0,0.1,...,1},
        minor ytick={0,0.025,...,1},
        major grid style={white!20!black},
        minor grid style={white!70!black},
        xlabel shift={-2pt},
        ylabel shift={-3pt},
        xmode=log]
        \addplot+[black,solid,mark=none, ultra thick]                            table[x=ncands,y=rec_at_0.5] {data/obj_cands/SBD_val_MCG_boxes.txt};
        \addplot+[red,solid,mark=none, thick]                                    table[x=ncands,y=rec_at_0.5] {data/obj_cands/SBD_val_SCG_boxes.txt};
        \addplot+[only marks,blue,solid,mark=triangle,mark size=1.9, thick]      table[x=ncands,y=rec_at_0.5] {data/obj_cands/SBD_val_GOP_boxes.txt};
	  	\addplot+[only marks,cyan,solid,mark=square,mark size=1.5, thick]        table[x=ncands,y=rec_at_0.5] {data/obj_cands/SBD_val_GLS_boxes.txt};
		\addplot+[only marks,red,solid,mark=asterisk, mark size=1.75, thick]     table[x=ncands,y=rec_at_0.5] {data/obj_cands/SBD_val_RIGOR_boxes.txt};
   	    \addplot+[blue,only marks,solid,mark=o, mark size=1.6, thick]            table[x=ncands,y=rec_at_0.5] {data/obj_cands/SBD_val_RP_boxes.txt};
		\addplot+[only marks,black,solid,mark=square, mark size=1.5, thick]      table[x=ncands,y=rec_at_0.5] {data/obj_cands/SBD_val_SeSe_boxes.txt};
		\addplot+[only marks,olive  ,solid,mark=asterisk, mark size=1.75, thick] table[x=ncands,y=rec_at_0.5] {data/obj_cands/SBD_val_EB_boxes.txt};
    	\addplot+[only marks,red    ,solid,mark=+, mark size=1.75, thick]        table[x=ncands,y=rec_at_0.5] {data/obj_cands/SBD_val_BING_boxes.txt};
	   \addplot+[only marks,black  ,solid,mark=x, thick]                         table[x=ncands,y=rec_at_0.5] {data/obj_cands/SBD_val_Obj_boxes.txt};
	    \addplot+[black,dashed,mark=none,thick]                                  table[x=ncands,y=rec_at_0.5] {data/obj_cands/SBD_val_QT_boxes.txt};

		\addplot+[black,solid,mark=none, ultra thick]                            table[x=ncands,y=rec_at_0.7] {data/obj_cands/SBD_val_MCG_boxes.txt};
        \addplot+[red,solid,mark=none, thick]                                    table[x=ncands,y=rec_at_0.7] {data/obj_cands/SBD_val_SCG_boxes.txt};
        \addplot+[only marks,blue,solid,mark=triangle,mark size=1.9, thick]      table[x=ncands,y=rec_at_0.7] {data/obj_cands/SBD_val_GOP_boxes.txt};
	  	\addplot+[only marks,cyan,solid,mark=square,mark size=1.5, thick]        table[x=ncands,y=rec_at_0.7] {data/obj_cands/SBD_val_GLS_boxes.txt};
		\addplot+[only marks,red,solid,mark=asterisk, mark size=1.75, thick]     table[x=ncands,y=rec_at_0.7] {data/obj_cands/SBD_val_RIGOR_boxes.txt};
   	    \addplot+[blue,only marks,solid,mark size=1.6,mark=o, thick]             table[x=ncands,y=rec_at_0.7] {data/obj_cands/SBD_val_RP_boxes.txt};
		\addplot+[only marks,black,solid,mark=square, mark size=1.5, thick]      table[x=ncands,y=rec_at_0.7] {data/obj_cands/SBD_val_SeSe_boxes.txt};
	    \addplot+[only marks,olive  ,solid,mark=asterisk, mark size=1.75, thick] table[x=ncands,y=rec_at_0.7] {data/obj_cands/SBD_val_EB_boxes.txt};
		\addplot+[only marks,red    ,solid,mark=+, mark size=1.75, thick]        table[x=ncands,y=rec_at_0.7] {data/obj_cands/SBD_val_BING_boxes.txt};
	    \addplot+[only marks,black  ,solid,mark=x, thick]                        table[x=ncands,y=rec_at_0.7] {data/obj_cands/SBD_val_Obj_boxes.txt};
        \addplot+[black,dashed,mark=none,thick]                                  table[x=ncands,y=rec_at_0.7] {data/obj_cands/SBD_val_QT_boxes.txt};

        \addplot+[black,solid,mark=none, ultra thick]                            table[x=ncands,y=rec_at_0.85] {data/obj_cands/SBD_val_MCG_boxes.txt};
        \addplot+[red,solid,mark=none, thick]                                    table[x=ncands,y=rec_at_0.85] {data/obj_cands/SBD_val_SCG_boxes.txt};
        \addplot+[only marks,blue,solid,mark=triangle,mark size=1.9, thick]      table[x=ncands,y=rec_at_0.85] {data/obj_cands/SBD_val_GOP_boxes.txt};
	  	\addplot+[only marks,cyan,solid,mark=square,mark size=1.5, thick]        table[x=ncands,y=rec_at_0.85] {data/obj_cands/SBD_val_GLS_boxes.txt};
		\addplot+[only marks,red,solid,mark=asterisk, mark size=1.75, thick]     table[x=ncands,y=rec_at_0.85] {data/obj_cands/SBD_val_RIGOR_boxes.txt};
   	    \addplot+[blue,only marks,solid,mark=o, mark size=1.6, thick]            table[x=ncands,y=rec_at_0.85] {data/obj_cands/SBD_val_RP_boxes.txt};
		\addplot+[only marks,black,solid,mark=square, mark size=1.5, thick]      table[x=ncands,y=rec_at_0.85] {data/obj_cands/SBD_val_SeSe_boxes.txt};
		\addplot+[only marks,olive  ,solid,mark=asterisk, mark size=1.75, thick] table[x=ncands,y=rec_at_0.85] {data/obj_cands/SBD_val_EB_boxes.txt};
		\addplot+[only marks,red    ,solid,mark=+, mark size=1.75, thick]        table[x=ncands,y=rec_at_0.85] {data/obj_cands/SBD_val_BING_boxes.txt};
	    \addplot+[only marks,black  ,solid,mark=x, thick]                        table[x=ncands,y=rec_at_0.85] {data/obj_cands/SBD_val_Obj_boxes.txt};
        \addplot+[black,dashed,mark=none,thick]                                  table[x=ncands,y=rec_at_0.85] {data/obj_cands/SBD_val_QT_boxes.txt};

	   \node[draw,fill=white,solid,anchor=north west] at (axis cs:20,0.58) {$J=0.5$};
	   \node[draw,fill=white,solid,anchor=north west] at (axis cs:20,0.38) {$J=0.7$};
	   \node[draw,fill=white,solid,anchor=north west] at (axis cs:20,0.2 ) {$J=0.85$};
	\end{axis}
   \end{tikzpicture}\end{minipage}
\begin{minipage}[b]{0.39\linewidth}
\centering
\scriptsize \hspace{5mm}COCO\\[1mm]
\begin{tikzpicture}[/pgfplots/width=1.1\linewidth, /pgfplots/height=1.27\linewidth]
    \begin{axis}[ymin=0,ymax=1,xmin=20,xmax=40000,enlargelimits=false,
        xlabel=Number of proposals,
        ylabel=Recall,
         grid=both,
        grid style=dotted,
        axis equal image=false,
        legend pos=outer north east,
        ytick={0,0.1,...,1},
        minor ytick={0,0.025,...,1},
        major grid style={white!20!black},
        minor grid style={white!70!black},
        xlabel shift={-2pt},
        ylabel shift={-3pt},
        xmode=log]
        \addplot[black,solid,mark=none, ultra thick]                            coordinates {( 1, 0.7)( 1, 0.8)};
        \addplot[red,solid,mark=none, thick]                                    coordinates {( 1, 0.7)( 1, 0.8)};
	    \addplot[cyan,solid,mark=none, thick]                                   coordinates {( 1, 0.7)( 1, 0.8)};
	    \addplot[olive,solid,mark=none, thick]                                  coordinates {( 1, 0.7)( 1, 0.8)};
        \addplot[only marks,blue   ,solid,mark=triangle,mark size=1.9, thick]   coordinates {( 1, 0.7)( 1, 0.8)};
	  	\addplot[only marks,cyan   ,solid,mark=square,mark size=1.5, thick]     coordinates {( 1, 0.7)( 1, 0.8)};
	    \addplot[only marks,red    ,solid,mark=asterisk, mark size=1.75, thick] coordinates {( 1, 0.7)( 1, 0.8)};
	    \addplot[only marks,magenta,solid,mark=triangle,mark size=1.9, thick]   coordinates {( 1, 0.7)( 1, 0.8)};
   	    \addplot[only marks,black  ,solid,mark=square, mark size=1.5, thick]    coordinates {( 1, 0.7)( 1, 0.8)};
	    \addplot[only marks,blue   ,solid,mark=o,mark size=1.6, thick]          coordinates {( 1, 0.7)( 1, 0.8)};
	    \addplot[only marks,olive  ,solid,mark=asterisk, mark size=1.75, thick] coordinates {( 1, 0.7)( 1, 0.8)};
	    \addplot[only marks,red    ,solid,mark=+, mark size=1.75, thick]        coordinates {( 1, 0.7)( 1, 0.8)};
	    \addplot[only marks,black  ,solid,mark=x, thick]                        coordinates {( 1, 0.7)( 1, 0.8)};
        \addplot[black,dashed,mark=none,thick]                                  coordinates {( 1, 0.7)( 1, 0.8)};

		\addlegendentry{MCG-Our}
        \addlegendentry{SCG-Our}
        \addlegendentry{CPMC~\cite{Carreira2012b}}
	    \addlegendentry{CI~\cite{Endres2014}}
        \addlegendentry{GOP~\cite{Kraehenbuehl2014}}
		\addlegendentry{GLS~\cite{Rantalankila2014}}
	   \addlegendentry{RIGOR~\cite{Humayun2014}}
		\addlegendentry{ShSh~\cite{Kim2012}}	   
		\addlegendentry{SeSe~\cite{Uijlings2013}}
	   \addlegendentry{RP~\cite{Manen2013}}
	   \addlegendentry{EB~\cite{Zitnick2014}}
	   \addlegendentry{BING~\cite{Cheng2014}}
	   \addlegendentry{Obj~\cite{Alexe2012}}
	   \addlegendentry{Quadtree}
	   
        \addplot+[black,solid,mark=none, ultra thick]                            table[x=ncands,y=rec_at_0.5] {data/obj_cands/COCO_val2014_MCG_boxes.txt};
        \addplot+[red,solid,mark=none, thick]                                    table[x=ncands,y=rec_at_0.5] {data/obj_cands/COCO_val2014_SCG_boxes.txt};
        \addplot+[only marks,blue,solid,mark=triangle,mark size=1.9, thick]      table[x=ncands,y=rec_at_0.5] {data/obj_cands/COCO_val2014_GOP_boxes.txt};
	  	\addplot+[only marks,cyan,solid,mark=square,mark size=1.5, thick]        table[x=ncands,y=rec_at_0.5] {data/obj_cands/COCO_val2014_GLS_boxes.txt};
	    \addplot+[only marks,red,solid,mark=asterisk, mark size=1.75, thick]     table[x=ncands,y=rec_at_0.5] {data/obj_cands/COCO_val2014_RIGOR_boxes.txt};
   	    \addplot+[blue,only marks,solid,mark=o, mark size=1.6, thick]            table[x=ncands,y=rec_at_0.5] {data/obj_cands/COCO_val2014_RP_boxes.txt};
	    \addplot+[only marks,black,solid,mark=square, mark size=1.5, thick]      table[x=ncands,y=rec_at_0.5] {data/obj_cands/COCO_val2014_SeSe_boxes.txt};
		\addplot+[only marks,olive  ,solid,mark=asterisk, mark size=1.75, thick] table[x=ncands,y=rec_at_0.5] {data/obj_cands/COCO_val2014_EB_boxes.txt};
		\addplot+[only marks,red    ,solid,mark=+, mark size=1.75, thick]        table[x=ncands,y=rec_at_0.5] {data/obj_cands/COCO_val2014_BING_boxes.txt};
	    \addplot+[only marks,black  ,solid,mark=x, thick]                        table[x=ncands,y=rec_at_0.5] {data/obj_cands/COCO_val2014_Obj_boxes.txt};
        \addplot+[black,dashed,mark=none,thick]                                  table[x=ncands,y=rec_at_0.5] {data/obj_cands/COCO_val2014_QT_boxes.txt};

		\addplot+[black,solid,mark=none, ultra thick]                            table[x=ncands,y=rec_at_0.7] {data/obj_cands/COCO_val2014_MCG_boxes.txt};
        \addplot+[red,solid,mark=none, thick]                                    table[x=ncands,y=rec_at_0.7] {data/obj_cands/COCO_val2014_SCG_boxes.txt};
        \addplot+[only marks,blue,solid,mark=triangle,mark size=1.9, thick]      table[x=ncands,y=rec_at_0.7] {data/obj_cands/COCO_val2014_GOP_boxes.txt};
	  	\addplot+[only marks,cyan,solid,mark=square,mark size=1.5, thick]        table[x=ncands,y=rec_at_0.7] {data/obj_cands/COCO_val2014_GLS_boxes.txt};
	    \addplot+[only marks,red,solid,mark=asterisk, mark size=1.75, thick]     table[x=ncands,y=rec_at_0.7] {data/obj_cands/COCO_val2014_RIGOR_boxes.txt};
   	    \addplot+[blue,only marks,solid,mark=o, mark size=1.6, thick]            table[x=ncands,y=rec_at_0.7] {data/obj_cands/COCO_val2014_RP_boxes.txt};
	    \addplot+[only marks,black,solid,mark=square, mark size=1.5, thick]      table[x=ncands,y=rec_at_0.7] {data/obj_cands/COCO_val2014_SeSe_boxes.txt};
		\addplot+[only marks,olive  ,solid,mark=asterisk, mark size=1.75, thick] table[x=ncands,y=rec_at_0.7] {data/obj_cands/COCO_val2014_EB_boxes.txt};
		\addplot+[only marks,red    ,solid,mark=+, mark size=1.75, thick]        table[x=ncands,y=rec_at_0.7] {data/obj_cands/COCO_val2014_BING_boxes.txt};
	    \addplot+[only marks,black  ,solid,mark=x, thick]                        table[x=ncands,y=rec_at_0.7] {data/obj_cands/COCO_val2014_Obj_boxes.txt};
        \addplot+[black,dashed,mark=none,thick]                                  table[x=ncands,y=rec_at_0.7] {data/obj_cands/COCO_val2014_QT_boxes.txt};

        \addplot+[black,solid,mark=none, ultra thick]                            table[x=ncands,y=rec_at_0.85] {data/obj_cands/COCO_val2014_MCG_boxes.txt};
        \addplot+[red,solid,mark=none, thick]                                    table[x=ncands,y=rec_at_0.85] {data/obj_cands/COCO_val2014_SCG_boxes.txt};
        \addplot+[only marks,blue,solid,mark=triangle,mark size=1.9, thick]      table[x=ncands,y=rec_at_0.85] {data/obj_cands/COCO_val2014_GOP_boxes.txt};
	  	\addplot+[only marks,cyan,solid,mark=square,mark size=1.5, thick]        table[x=ncands,y=rec_at_0.85] {data/obj_cands/COCO_val2014_GLS_boxes.txt};
	    \addplot+[only marks,red,solid,mark=asterisk, mark size=1.75, thick]     table[x=ncands,y=rec_at_0.85] {data/obj_cands/COCO_val2014_RIGOR_boxes.txt};
   	    \addplot+[blue,only marks,solid,mark=o, mark size=1.6,thick]             table[x=ncands,y=rec_at_0.85] {data/obj_cands/COCO_val2014_RP_boxes.txt};
	    \addplot+[only marks,black,solid,mark=square, mark size=1.5, thick]      table[x=ncands,y=rec_at_0.85] {data/obj_cands/COCO_val2014_SeSe_boxes.txt};
	    \addplot+[only marks,olive  ,solid,mark=asterisk, mark size=1.75, thick] table[x=ncands,y=rec_at_0.85] {data/obj_cands/COCO_val2014_EB_boxes.txt};
	    \addplot+[only marks,red    ,solid,mark=+, mark size=1.75, thick]        table[x=ncands,y=rec_at_0.85] {data/obj_cands/COCO_val2014_BING_boxes.txt};
	    \addplot+[only marks,black  ,solid,mark=x, thick]                        table[x=ncands,y=rec_at_0.85] {data/obj_cands/COCO_val2014_Obj_boxes.txt};
        \addplot+[black,dashed,mark=none,thick]                                  table[x=ncands,y=rec_at_0.85] {data/obj_cands/COCO_val2014_QT_boxes.txt};

	   \node[draw,fill=white,solid,anchor=north west] at (axis cs:20,0.28) {$J=0.5$};
	   \node[draw,fill=white,solid,anchor=north west] at (axis cs:20,0.16) {$J=0.7$};
	   \node[draw,fill=white,solid,anchor=north west] at (axis cs:20,0.08) {$J=0.85$};
	\end{axis}
   \end{tikzpicture}\end{minipage}}
   \caption{\textbf{Bounding-Box Proposals: Recall at different Jaccard levels.} Percentage of annotated objects for which there is a bounding box proposal whose overlap with the ground-truth boxes is above $J=0.5$, $J=0.7$, and $J=0.85$, for different number of proposals per image. Results on SegVOC12, SBD, and COCO.}
   \label{fig:recall_boxes}
\end{figure*}

As commented on the measures description, $J_i$ shows mean aggregate results, so they can mask
the distribution of quality among objects in the database.
Figure~\ref{fig:recall_segmented} shows the recall at three different Jaccard levels.
First, these plots further highlight how challenging COCO is, since we observe a significant drop in performance,
more pronounced than when measured by $J_i$ and $J_c$.
Another interesting result comes from observing the evolution of the plots for the three different Jaccard values.
Take for instance the performance of GOP~(\ref{fig:recall:gop}) against MCG-Our (\ref{fig:recall:mcg}) in SBD.
While for $J\!=\!0.5$ GOP slightly outperforms MCG, the higher the threshold, the better MCG.
Overall, MCG has specially good results at higher $J$ values.
In other words, if one looks for proposals of very high accuracy, MCG is the method with highest recall,
at all regimes and in all databases.
In all measures and databases, SCG (\ref{fig:recall:scg}) obtains very competitive results, especially if we take into account that it is 7$\times$ faster than MCG,
as we will see in next sections.

The complementarity of MCG with respect to other proposal techniques, and their combination using the Pareto front optimization is studied in~\cite{Pont-Tuset2015b}.

\paragraph*{\textbf{Boxes Proposals: Comparison with State of the Art}}
Although focused in providing segmented object proposals, MCG may also be used to provide boxes proposals, by taking the bounding 
box of the segmented proposals and deduplicating them.
We add the state of the art in boxes proposals~\cite{Zitnick2014}, \cite{Cheng2014}, \cite{Manen2013}, and~\cite{Alexe2012} to the comparison.
Figure~\ref{fig:recall_boxes} shows the recall values of the boxes results when compared to the bounding boxes of the annotated objects, for three different
Jaccard thresholds.

While many of the techniques specifically tailored to boxes proposals are competitive at $J\!=\!0.5$, their performance drops 
significantly at higher Jaccard thresholds.
Despite being tailored to segmented proposals, MCG clearly outperforms the state of the art if you look for
precise localization of the bounding boxes.
Again, SCG is very competitive, especially taking its speed into account.

\paragraph*{\textbf{MCG and SCG Time per Image}}
Table~\ref{timing} shows the time per image of the main steps of our approach,
from the raw image to the contours, the hierarchical segmentation, and the proposal generation.
All times are computed using a single core on a Linux machine.
Our full MCG takes about 25 s. per image to compute the multiscale hierarchies, and 17 s. to generate and
rank the $5\,038$ proposals on the validation set of SegVOC12.
Our single-scale SCG produces a segmentation hierarchy of better quality than gPb-ucm~\cite{Arbelaez2011} in less than 3
seconds and with significant less memory footprint.

\begin{figure*}
\setlength{\fboxsep}{0pt}
\fbox{\includegraphics[width=0.19\linewidth]{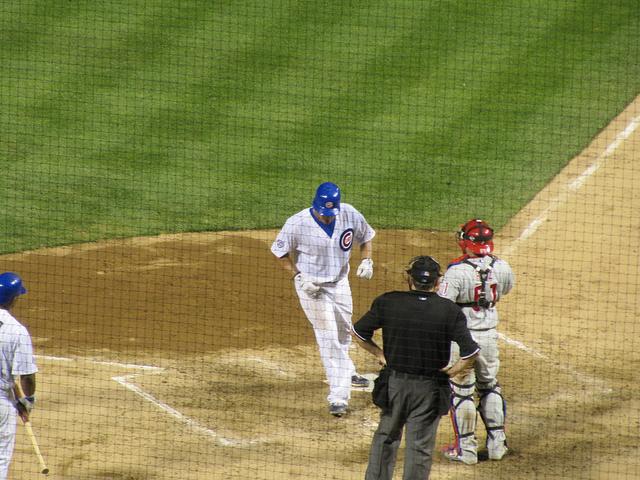}}
\hfill
\fbox{\includegraphics[width=0.19\linewidth]{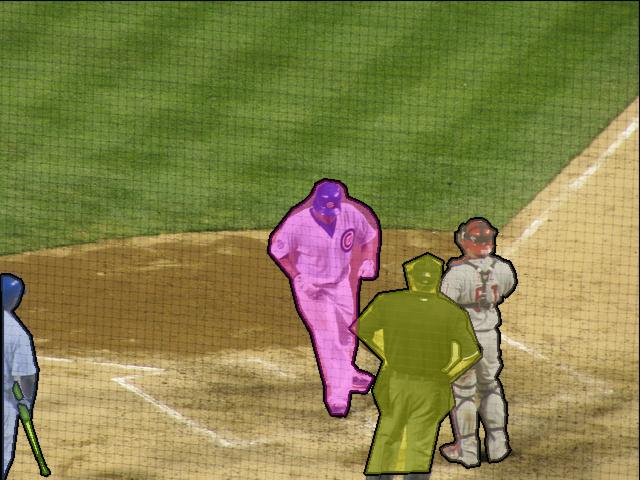}}
\hfill
\fbox{\includegraphics[width=0.19\linewidth]{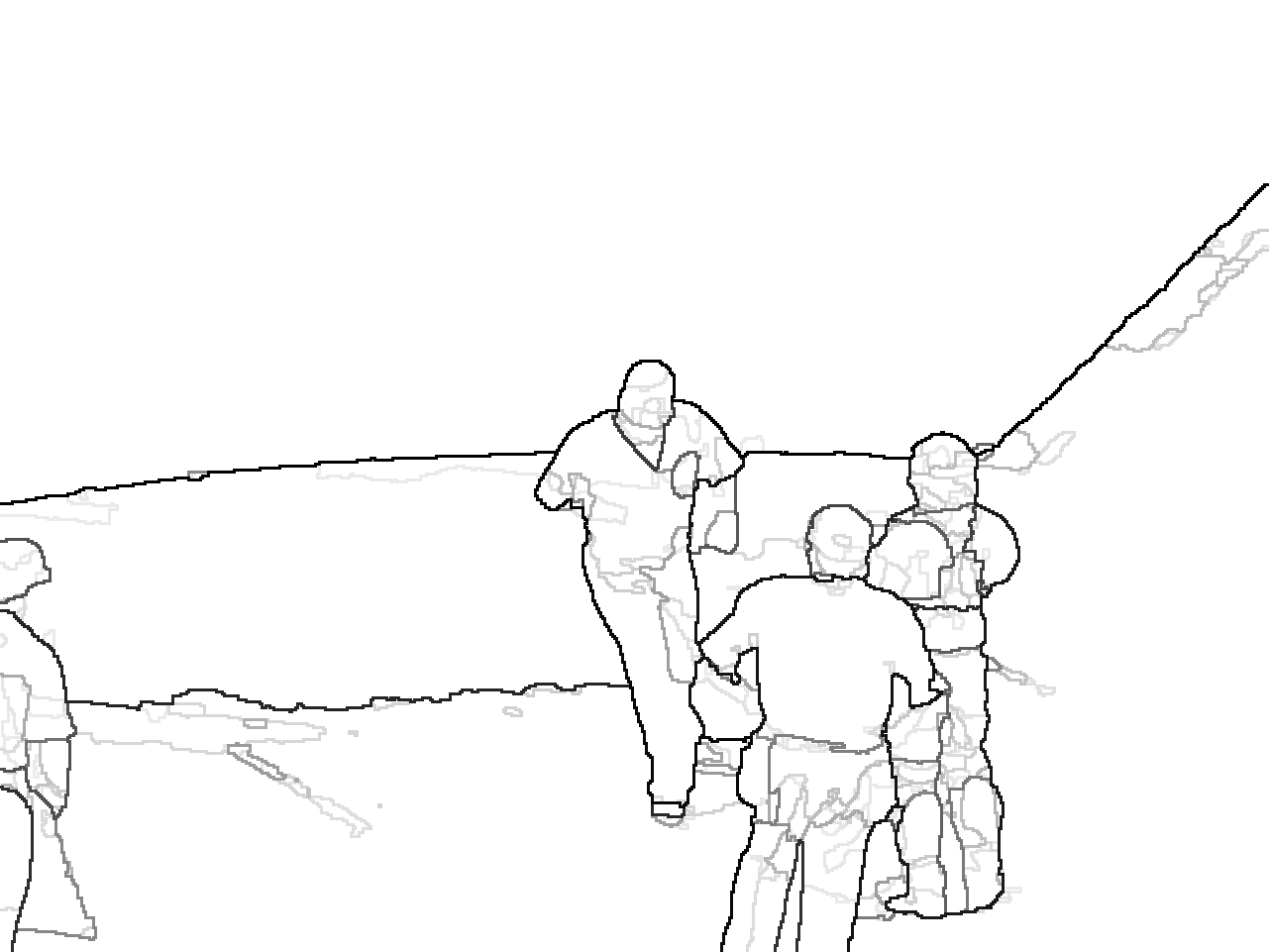}}
\hfill
\fbox{\includegraphics[width=0.127\linewidth]{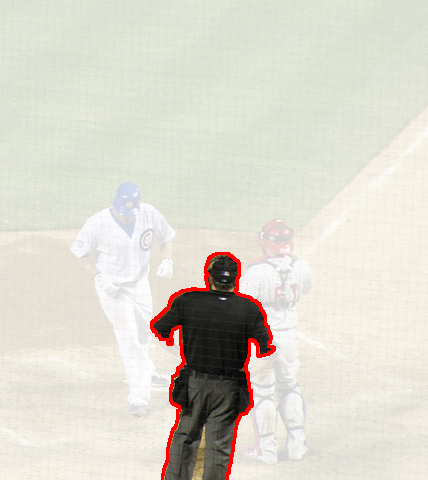}}
\hfill
\fbox{\includegraphics[width=0.127\linewidth]{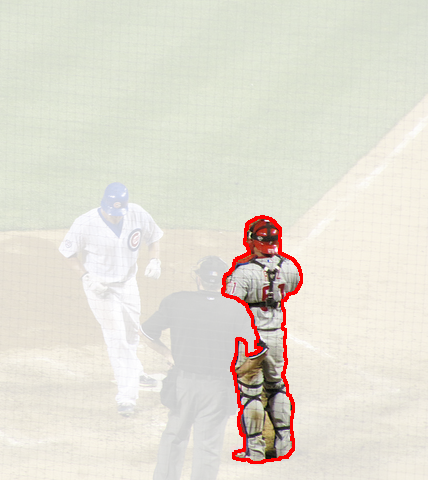}}
\hfill
\fbox{\includegraphics[width=0.127\linewidth]{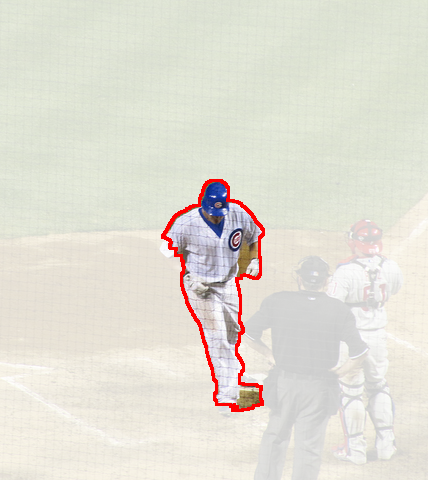}}
\hfill\\[1mm]
\fbox{\includegraphics[width=0.19\linewidth]{figures/examples/COCO_val2014_000000000136.jpg}}
\hfill
\fbox{\includegraphics[width=0.19\linewidth]{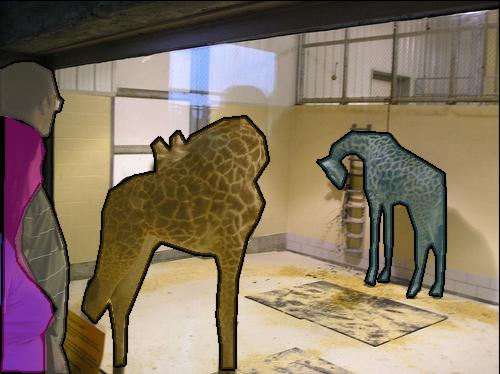}}
\hfill
\fbox{\includegraphics[width=0.19\linewidth]{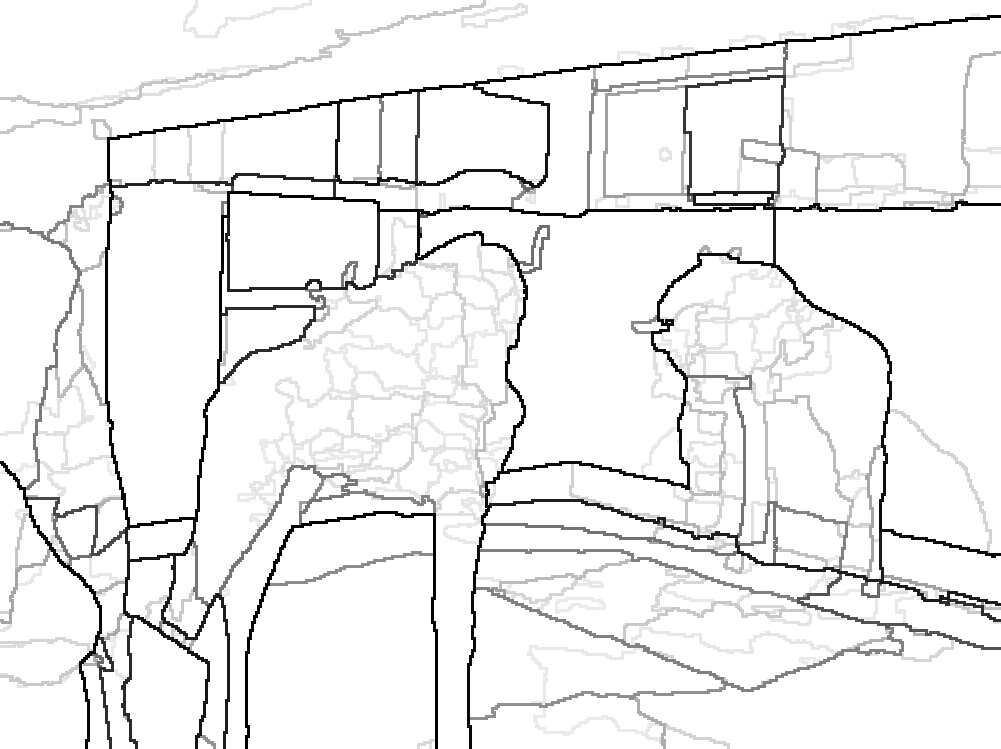}}
\hfill
\fbox{\includegraphics[width=0.127\linewidth]{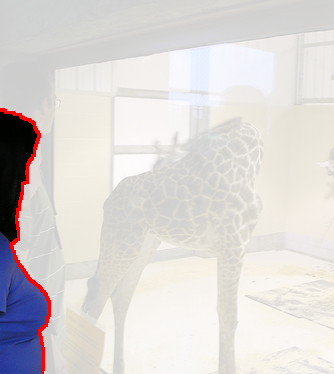}}
\hfill
\fbox{\includegraphics[width=0.127\linewidth]{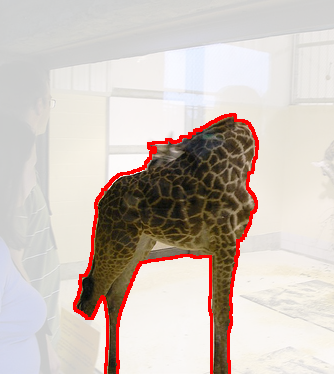}}
\hfill
\fbox{\includegraphics[width=0.127\linewidth]{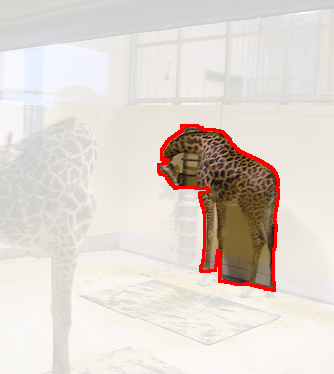}}
\hfill\\[1mm]
\fbox{\includegraphics[width=0.19\linewidth]{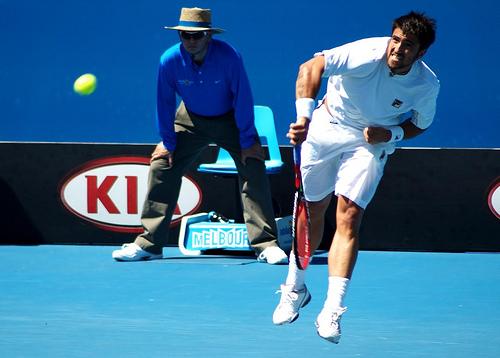}}
\hfill
\fbox{\includegraphics[width=0.19\linewidth]{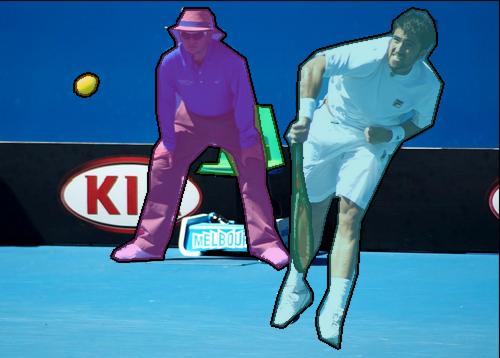}}
\hfill
\fbox{\includegraphics[width=0.19\linewidth]{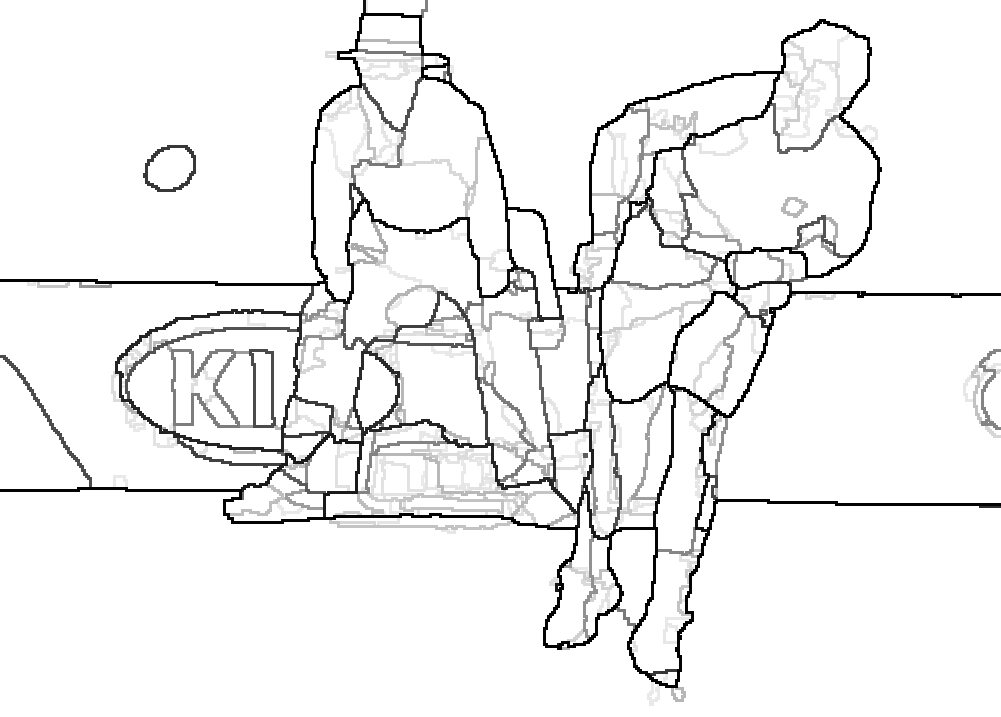}}
\hfill
\fbox{\includegraphics[width=0.127\linewidth]{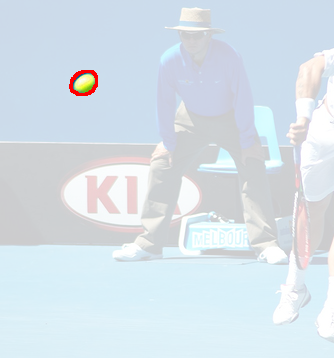}}
\hfill
\fbox{\includegraphics[width=0.127\linewidth]{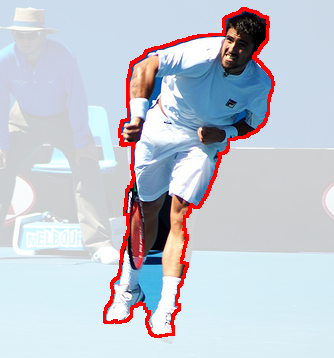}}
\hfill
\fbox{\includegraphics[width=0.127\linewidth]{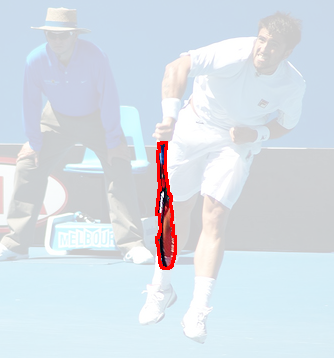}}
\hfill\\[1mm]
\fbox{\includegraphics[width=0.19\linewidth]{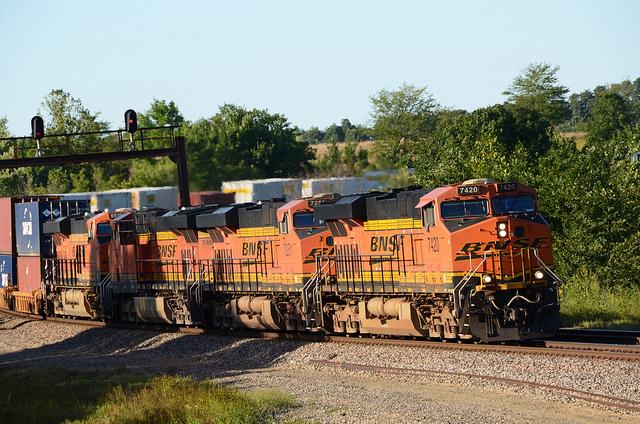}}
\hfill
\fbox{\includegraphics[width=0.19\linewidth]{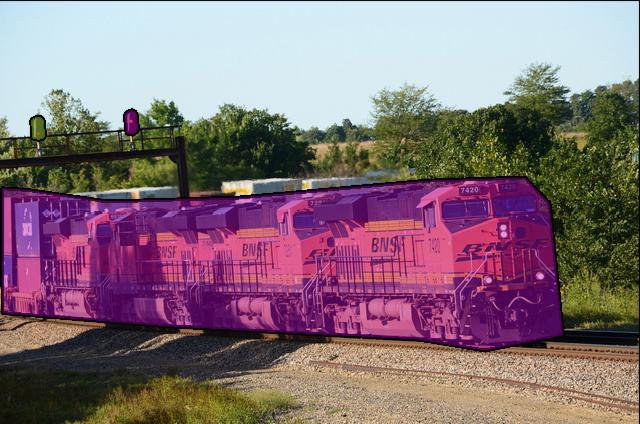}}
\hfill
\fbox{\includegraphics[width=0.19\linewidth]{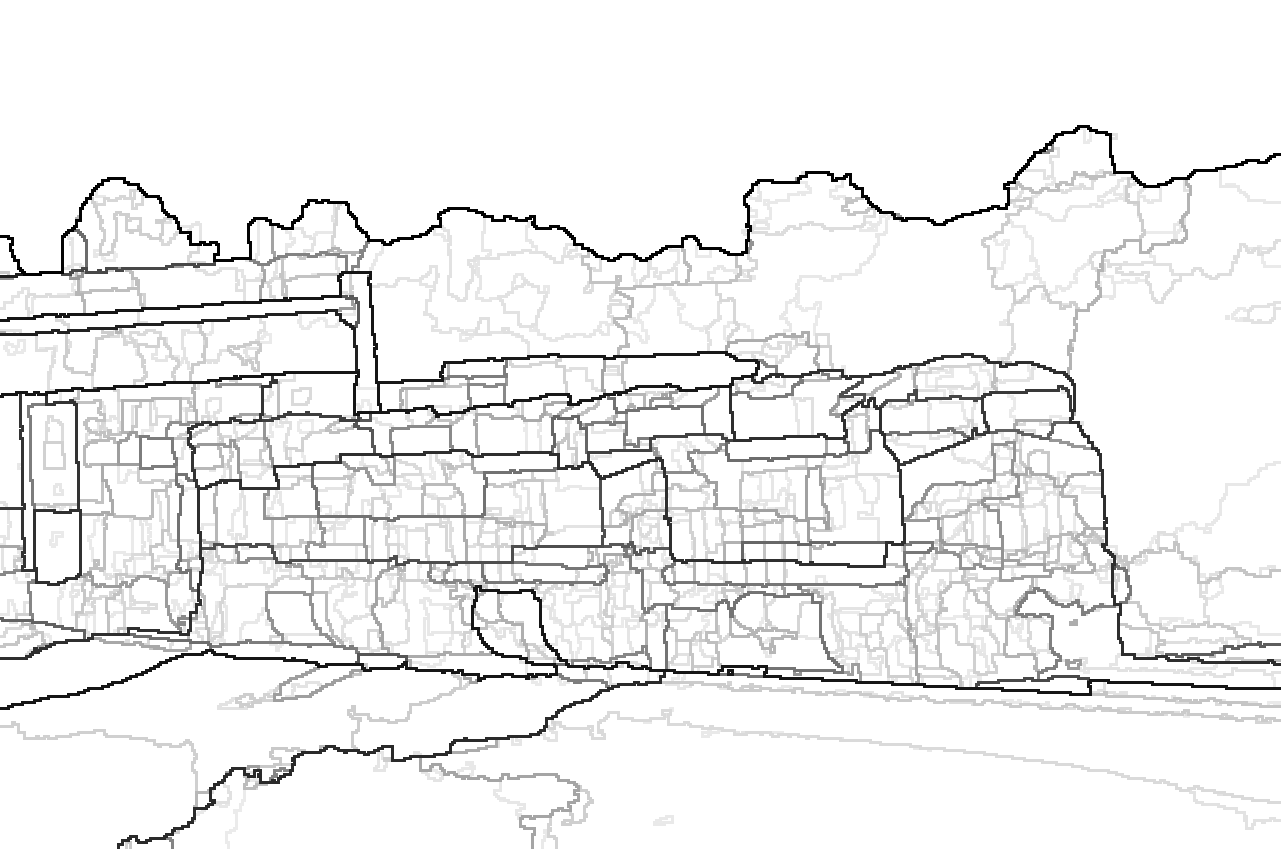}}
\hfill
\fbox{\includegraphics[width=0.19\linewidth]{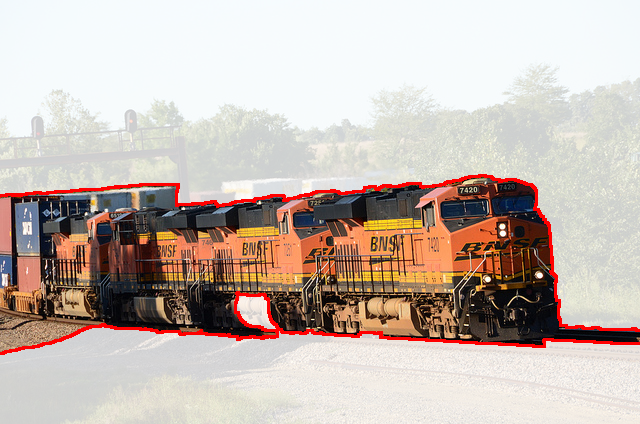}}
\hfill
\fbox{\includegraphics[width=0.095\linewidth]{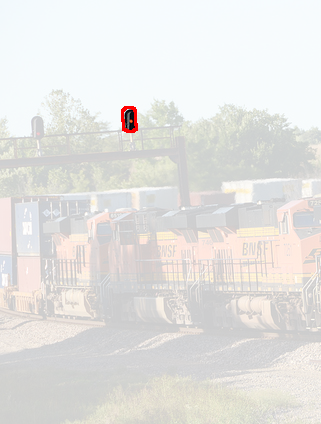}}
\hfill
\fbox{\includegraphics[width=0.095\linewidth]{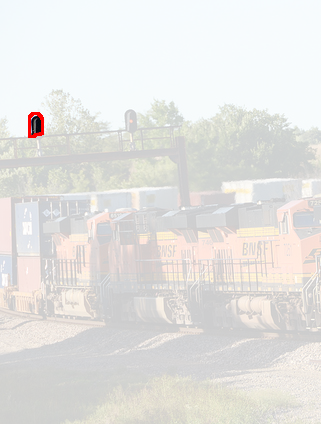}}
\hfill
\caption{\textbf{COCO Qualitative Results}: Image, ground truth, multi-scale UCM and best MCG proposals among the 500 best ranked. (More qualitative examples in the supplemental material.)}
\label{fig:qualitative_examples}
\end{figure*}

\begin{table}[h]
\centering
\scalebox{0.95}{\begin{tabular}{l|ccc|c}
\cmidrule[\heavyrulewidth](l{-2pt}){2-5}
\multicolumn{1}{c}{}    & Contour   & Hierarchical & Candidate  & \multirow{2}{*}{Total}\\
\multicolumn{1}{c}{}    & Detection & Segmentation & Generation &                       \\
    \midrule
MCG                     &    4.6$\,\pm\,$1.3 &    20.5$\,\pm\,$5.6       &   17.0$\,\pm\,$9.8     &     42.2$\,\pm\,$14.8  \\
SCG                     &    1.0$\,\pm\,$0.3&     \,\,\,2.7$\,\pm\,$0.7 &   \,\,\,2.6$\,\pm\,$0.4     &      \,\,\,6.2$\,\pm\,\,\,\,$1.1  \\
 \bottomrule
\end{tabular}}
\vspace{1mm}
\caption{Time in seconds per image of MCG and SCG}
\label{timing}
\end{table}

Table~\ref{timing_soa} shows the time-per-image results compared to the rest of state of the art in segmented proposals generation,
all run on the same single-core Linux machine.

\begin{table}[h]
\centering
\scalebox{0.95}{\begin{tabular}{l|c}
\cmidrule[\heavyrulewidth](l{-2pt}){2-2}
\multicolumn{1}{c}{}    &  Proposal  \\
\multicolumn{1}{c}{}    &  Generation \\
    \midrule
MCG (Our) &    \,\,\,42.2$\,\pm\,$14.8  \\
SCG (Our) &     \,\,\,\,\,\,6.2$\,\pm\,\,\,\,$1.1 \\
\midrule
GOP~\cite{Kraehenbuehl2014}&     \,\,\,\,\,\,1.0$\,\pm\,\,\,\,$0.3 \\
GLS~\cite{Rantalankila2014}&    \,\,\,\,\,\,7.9$\,\pm\,\,\,\,$1.7 \\
SeSe~\cite{Uijlings2013}&     \,\,\,15.9$\,\pm\,\,\,\,$5.2 \\
RIGOR~\cite{Humayun2014}&     \,\,\,31.6$\,\pm\,$16.0 \\
CPMC~\cite{Carreira2012b}&    $\geq$120 \\
CI~\cite{Endres2014}&     $\geq$120 \\
ShSh~\cite{Kim2012}&     $\geq$120 \\
\bottomrule
\end{tabular}}
\vspace{1mm}
\caption{Time comparison for all considered state-of-the-art techniques that produce segmented proposals. All run on the same single-core Linux machine.}
\label{timing_soa}
\end{table}

\paragraph*{\textbf{Practical considerations}}
One of the key aspects of object proposals is the size of the pool they generate. 
Depending on the application, one may need more precision and thus a bigger pool,
or one might need speed and thus a small pool in exchange for some loss of quality.
MCG and SCG provide a \textit{ranked} set of around
5\,000 and 2\,000 proposals, respectively, and one can take the $N$ first
in case the specific application needs a smaller pool.
From a practical point of view, this means that one does not need to re-parameterize them for the specific
needs of a certain application.

In contrast, the techniques that do not provide a ranking of the proposals,
need to be re-parameterized to adapt them to a different number of proposals, which 
is not desirable in practice.

On top of that, the results show that MCG and SCG have outstanding generalization power to unseen images
(recall that the results for SBD and COCO have been obtained using the learnt parameters on SegVOC12),
meaning that MCG and SCG offer the best chance to obtain competitive results in an unseen database without need to re-train.

Figure~\ref{fig:qualitative_examples} shows some qualitative results on COCO.

\section{Conclusions}
\label{sec:conclu}
We proposed Multiscale Combinatorial Grouping (MCG), a unified approach for bottom-up segmentation and
object proposal generation.
Our approach produces state-of-the-art contours, hierarchical regions, and object proposals. 
At its core are a fast eigenvector computation for normalized-cut segmentation and an efficient algorithm for combinatorial merging of hierarchical regions.
We also present Single-scale Combinatorial Grouping (SCG), a speeded up version of our technique that produces competitive results in under five seconds per image. 

We perform an extensive validation in BSDS500, SegVOC12, SBD, and COCO, showing the quality, robustness and scalability of MCG.
Recently, an independent study~\cite{schiele:bmvc14, Hosang2015} provided further evidence to the interest of MCG among the current state-of-the-art in object proposal generation. 
Moreover, our object candidates have already been employed as integral part of high performing recognition systems \cite{BharathECCV2014}.

In order to promote reproducible research on perceptual grouping,
all the resources of this project -- code, pre-computed results, and evaluation protocols -- are publicly available\footnote{\scriptsize{www.eecs.berkeley.edu/Research/Projects/CS/vision/grouping/mcg/}}.\vspace{4mm}

\paragraph*{\textbf{Acknowledgements}}
The last iterations of this work have been done while Jordi Pont-Tuset has been
at Prof. Luc Van Gool's Computer Vision Lab (CVL) of ETHZ, Switzerland.
This work has been partly developed in the framework of the project BIGGRAPH-TEC2013-43935-R and the FPU grant AP2008-01164; financed by the \textit{Spanish Ministerio de Econom\'ia y 
Competitividad}, and the European Regional Development Fund (ERDF).
This work was partially supported by ONR MURI N000141010933.

\bibliographystyle{IEEEtran}
\bibliography{MCG-arXiv2016}

\begin{IEEEbiography}[{\includegraphics[width=1in,height=1.25in,clip,keepaspectratio]{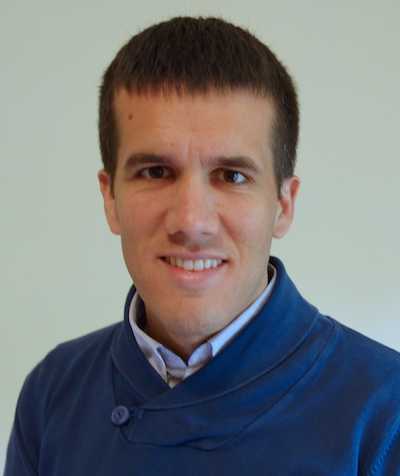}}]
{Jordi Pont-Tuset} is a post-doctoral researcher at ETHZ, Switzerland,
in Prof. Luc Van Gool's computer vision group (2015).
He received the degree in Mathematics in 2008, the degree in Electrical Engineering 
in 2008, the M.Sc. in Research on Information and Communication Technologies in 2010, and the Ph.D with honors in
2014; all from the Universitat Polit\`{e}cnica de Catalunya, BarcelonaTech (UPC).
He worked at Disney Research, Z\"urich (2014).
\end{IEEEbiography}

\begin{IEEEbiography}[{\includegraphics[width=1in,height=1.25in,clip,keepaspectratio]{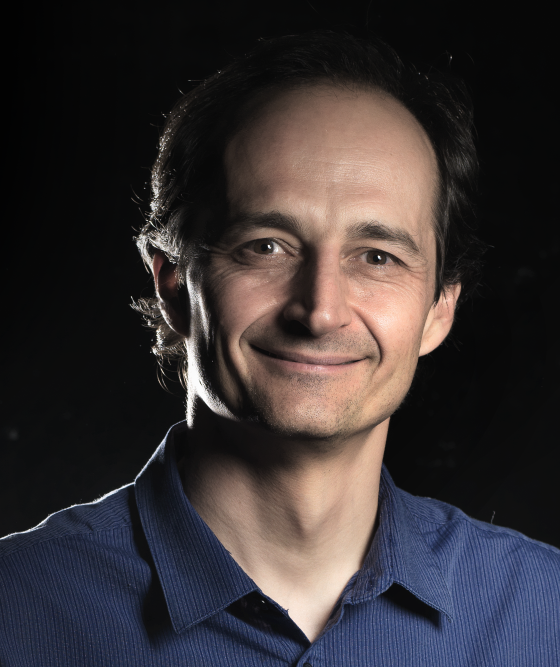}}]
{Pablo Arbel\'{a}ez} received a PhD with honors in Applied Mathematics from the Universit\'{e} Paris-Dauphine in 2005. 
He was a Research Scientist with the Computer Vision Group at UC Berkeley from 2007 to 2014. 
He currently holds a faculty position at Universidad de los Andes in Colombia.
His research interests are in computer vision, where he has worked on a number of problems, including perceptual grouping,
object recognition and the analysis of biomedical images.
\end{IEEEbiography}

\begin{IEEEbiography}[{\includegraphics[width=1in,height=1.25in,clip,keepaspectratio]{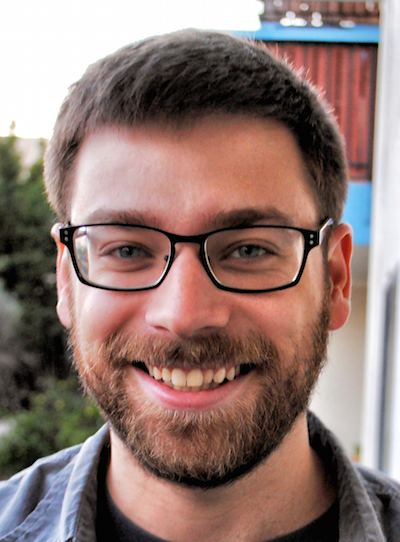}}]
{Jonathan T. Barron} is a senior research scientist at Google, working on computer vision and computational photography. He received a PhD in Computer Science from the University of California, Berkeley in 2013, where he was advised by Jitendra Malik, and he received a Honours BSc in Computer Science from the University of Toronto in 2007. His research interests include computer vision, machine learning, computational photography, shape reconstruction, and biological image analysis. He received a National Science Foundation Graduate Research Fellowship in 2009, and the C.V. Ramamoorthy Distinguished Research Award in 2013.
\end{IEEEbiography}

\begin{IEEEbiography}[{\includegraphics[width=1in,height=1.25in,clip,keepaspectratio]{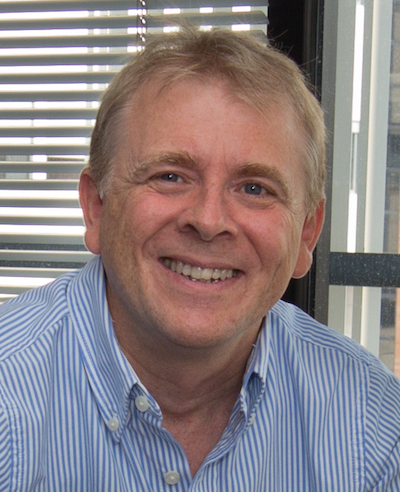}}]
{Ferran Marques} received the degree in Electrical Engineering and the Ph.D. from the Universitat Polit\`{e}cnica de Catalunya, BarcelonaTech (UPC),
where he is currently Professor at the department of Signal Theory and Communications.
In the term 2002-2004, he served as President of the European Association for Signal Processing (EURASIP).
He has served as Associate Editor of the IEEE Transactions on Image Processing 
(2009-2012) and as Area Editor for Signal Processing: Image 
Communication, Elsevier (2010-2014). In 2011, he received the 
Jaume Vicens Vives distinction for University Teaching Quality. 
Currently, he serves as Dean of the Electrical Engineering School 
(ETSETB-TelecomBCN) at UPC. He has published over 150 conference and 
journal papers, 2 books, and holds 4 international patents.  
\end{IEEEbiography}

\begin{IEEEbiography}[{\includegraphics[width=1in,height=1.25in,clip,keepaspectratio]{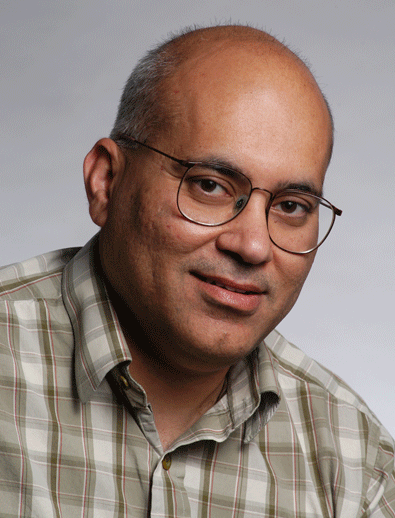}}]
{Jitendra Malik} is Arthur J. Chick Professor in the Department of Electrical Engineering and Computer Science at the University of California at Berkeley, where he also holds appointments in vision science and cognitive science. He received the PhD degree in Computer Science from Stanford University in 1985. In January 1986, he joined UC Berkeley as a faculty member in the EECS department where he served as Chair of the Computer Science Division during 2002-2006, and of the Department of EECS during 2004-2006. Jitendra Malik's group has worked on computer vision, computational modeling of biological vision, computer graphics and machine learning. Several well-known concepts and algorithms arose in this work, such as anisotropic diffusion, normalized cuts, high dynamic range imaging, shape contexts and poselets. According to Google Scholar, ten of his papers have received more than a thousand citations each. He has graduated 33 PhD students. Jitendra was awarded the Longuet-Higgins Award for ``A Contribution that has Stood the Test of Time'' twice, in 2007 and 2008. He is a Fellow of the IEEE and the ACM, a member of the National Academy of Engineering, and a fellow of the American Academy of  Arts and Sciences. He received the PAMI Distinguished Researcher Award in computer vision in 2013 and the K.S. Fu prize in 2014.
\end{IEEEbiography}

\end{document}